\documentclass[letterpaper]{article} 

\usepackage[utf8]{inputenc} 
\usepackage[T1]{fontenc}    

\usepackage{arxiv}

\usepackage{dutchcal}

\usepackage{natbib}

\setcitestyle{authoryear,open={(},close={)}} 

\usepackage{microtype}
\usepackage{float}
\usepackage{graphicx}
\usepackage{caption}
\usepackage{subfigure}
\usepackage{booktabs} 
\usepackage{subcaption}
\usepackage[table]{xcolor}
\usepackage{hhline}

\usepackage{amsmath}
\usepackage{amssymb}
\usepackage{mathtools}
\usepackage{amsthm}

\usepackage{enumitem}
\setlist[enumerate]{leftmargin=20pt}

\usepackage{hyperref}
\usepackage{bm}

\definecolor{RoyalBlue}{RGB}{0,0,255}
\hypersetup{
    colorlinks,
    linkcolor={cyan!90!black},
    citecolor={cyan!65!black},
    urlcolor={cyan!80!white}
}

\usepackage{url}
\usepackage{todonotes}

\usepackage{algorithm}
\usepackage{algpseudocode}

\usepackage{dsfont}
\usepackage{caption}
\usepackage{graphicx}
\usepackage{wrapfig}
\usepackage{graphics} 

\usepackage[capitalize,noabbrev]{cleveref}

\usepackage{xcolor}
\usepackage{tcolorbox}

\usepackage{multirow}
\usepackage{multicol}

\newtheorem{definition}{Definition}

\def \R {\mathbb{R}}

\def \P {\mathbb{P}}

\def \bx {{\bf x}}

\def \bz {{\bf z}}

\def \vx {{\bf x}}
\def \vt {{\bf t}}
\def \vv {{\bf v}}

\def \ff {{\mathcal{f} }}
\def \fg {{\mathcal{g} }}
\def \fh {{\mathcal{h} }}

\def \Dcal {{D_{\mathrm{cal}}}}
\def \Dth {{D_{\mathrm{th}}}}

\def \Dval {{D_{\mathrm{val}}}}

\def \Dcali {{D^{(i)}_{\mathrm{cal}}}}
\def \Dvali {{D^{(i)}_{\mathrm{val}}}}
\def \Dthi  {{D^{(i)}_{\mathrm{th}}}}
\def \Dthiy  {{D^{(i,y)}_{\mathrm{th}}}}

\def \bv {{\bf v}}

\def \yhat {{\hat{y}}}

\def \cE {\mathcal{E}}

\def \cG {\mathcal{G}}
\def \cH {\mathcal{H}}

\def \cO {\mathcal{O}}
\def \cP {\mathcal{P}}

\def \cX {\mathcal{X}}
\def \cY {\mathcal{Y}}

\renewcommand{\algorithmicrequire}{\textbf{Input:}}
\renewcommand{\algorithmicensure}{\textbf{Output:}}

\DeclareMathOperator*{\argmax}{arg\,max}
\DeclareMathOperator*{\argmin}{arg\,min}

\newcommand{\ldef}{\vcentcolon=}

\newcommand{\one}{\mathds{1}}

\newcommand{\mycomment}[1]{}

\def \err {\cE} 
\def \cov {\cP}

\def \haterr {\widehat{\cE}}
\def \hatcov {\widehat{\cP}}

\def \surrerr {\widetilde{\cE}}

\def \surrcov {\widetilde{\cP}}

\def \given {\mid }

\def \ourmethod {\texttt{Colander}}

\def \Dtraini {D_{\mathrm{train}}^{(i)}}

\def \Dval {D_{\mathrm{val}}}

\algnewcommand{\LineComment}[1]{ \State \(\color{gray}\triangleright \) {\color{gray}#1}}

\renewcommand{\algorithmiccomment}[1]{\hfill$\triangleright${\color{gray}{#1}}}

\title{\textbf{Pearls from Pebbles: Improved Confidence Functions for Auto-labeling}}

\author{{ Harit Vishwakarma} \\  \texttt{hvishwakarma@cs.wisc.edu} \\ \and \textbf{Reid (Yi) Chen} \\  \texttt{reid.chen@wisc.edu} \\  \and  
\textbf{\quad  Sui Jiet Tay} \\  \texttt{\quad  sstay2@wisc.edu} \\  \and  
\textbf{\qquad \quad Satya Sai Srinath Namburi} \\  \texttt{\qquad \quad sgnamburi@cs.wisc.edu} \\  \and  
\textbf{ Frederic Sala} \\ \texttt{ fredsala@cs.wisc.edu} \\ \and \textbf{Ramya Korlakai Vinayak} \\ \texttt{ramya@ece.wisc.edu} \\
\and
University of Wisconsin-Madison, WI, USA}
\date{}

\begin{document}

\maketitle



\begin{abstract}
Auto-labeling is an important family of techniques that produce labeled training sets with minimum manual labeling. A prominent variant, threshold-based auto-labeling (TBAL), works by finding a threshold on a model's confidence scores above which it can accurately label unlabeled data points. However, many models are known to produce overconfident scores, leading to poor TBAL  performance. While a natural idea is to apply off-the-shelf calibration methods to alleviate the overconfidence issue, such methods still fall short. Rather than experimenting with ad-hoc choices of confidence functions, we propose a framework for studying the \emph{optimal} TBAL confidence function. We develop a tractable version of the framework to obtain \texttt{Colander}  (Confidence functions for Efficient and Reliable Auto-labeling), a new post-hoc method specifically designed to maximize performance in TBAL systems. 
We perform an extensive empirical evaluation of our method \texttt{Colander}  and compare it against methods designed for calibration.
\texttt{Colander} achieves up to 60\% improvements on coverage over the baselines while maintaining auto-labeling error below $5\%$ and using the same amount of labeled data as the baselines. 
\end{abstract} 
 \section{Introduction}
\label{sec:intro}



\begin{wrapfigure}{r}
{0.5\textwidth}
    \centering
    \includegraphics[width=0.5\textwidth]{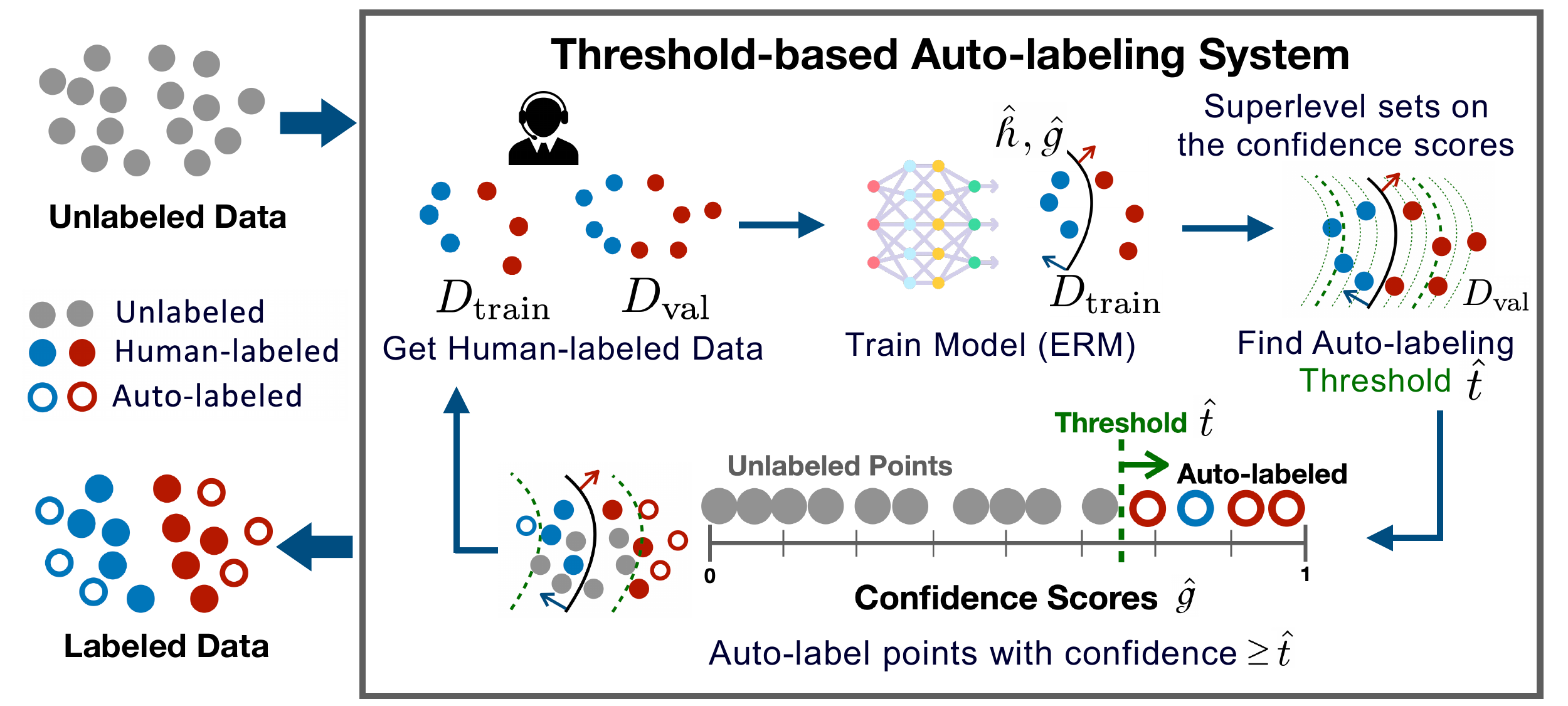}
    \vspace{-5pt}
    \caption{High-level diagram of an auto-labeling system. It takes unlabeled data as input and, with the help of expert labelers and ML models, outputs a labeled dataset.}
    \label{fig:auto-labeling-io}
    \vspace{-0pt}
\end{wrapfigure}

The demand for labeled data in machine learning (ML) is perpetual ~\citep{Fisher1936, deng2009imagenet, Touvron2023Llama2O}, yet obtaining such data is expensive and time-consuming, creating a bottleneck in ML workflows. Threshold-based auto-labeling (TBAL) emerges as a promising solution to obtain high-quality labeled data at low cost \citep{ sgt, qiu2020min-cost-al, vishwakarma2023promises}. A TBAL system (Figure \ref{fig:auto-labeling-io}) takes unlabeled data as input and outputs a labeled dataset. It works iteratively: in each iteration, it acquires human labels for a small chunk of data to train a model, then auto-labels points using the model's predictions where its \emph{confidence scores} are above a certain threshold. The threshold is determined using validation data so that the auto-labeled points meet a desired \emph{accuracy criteria}. The goal is to maximize \emph{coverage}---the fraction of auto-labeled points while maintaining the accuracy criteria.   
 
\begin{figure*}[t]
  \centering

  \mbox{

        \subfigure[ Softmax\label{fig:default_scores_dist}]{\includegraphics[width=0.198\linewidth]{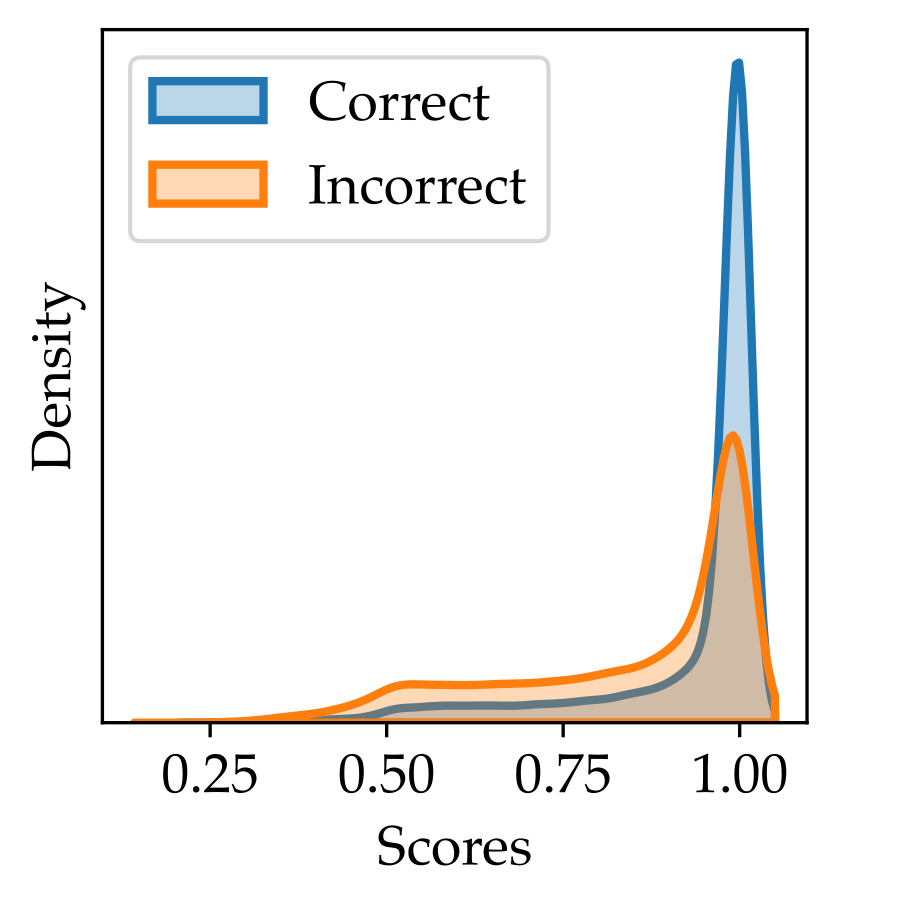}}
    
    \subfigure[ Temp. Scaling\label{fig:calib_scores}]{\includegraphics[width=0.198\linewidth]{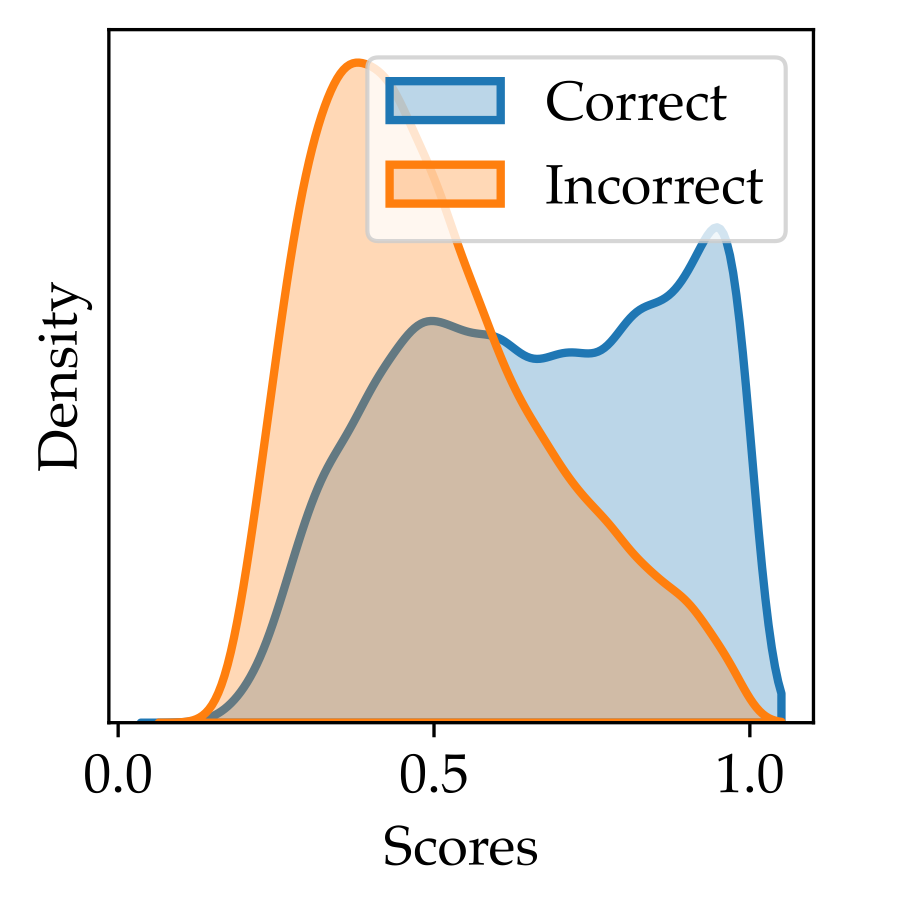}}

    \subfigure[ \ourmethod\, (Ours)\label{fig:colander_scores}]{\includegraphics[width=0.198\linewidth]{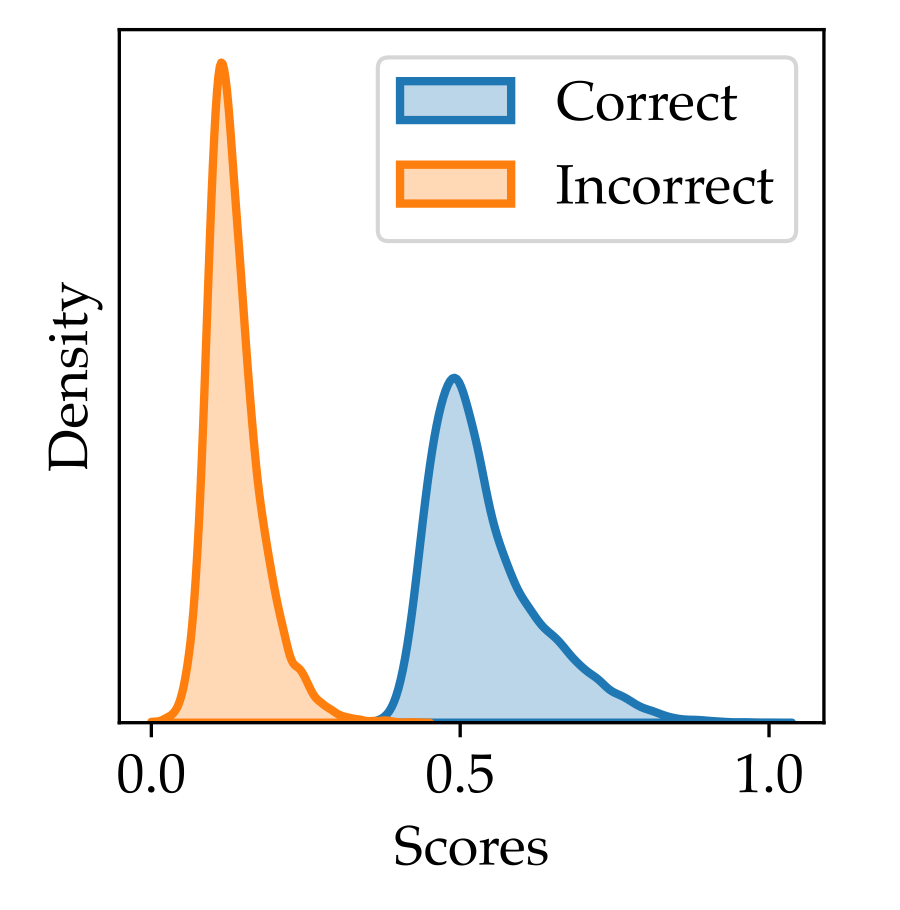}}

    \subfigure[ Coverage \label{fig:coverage}]{\includegraphics[width=0.195\linewidth]{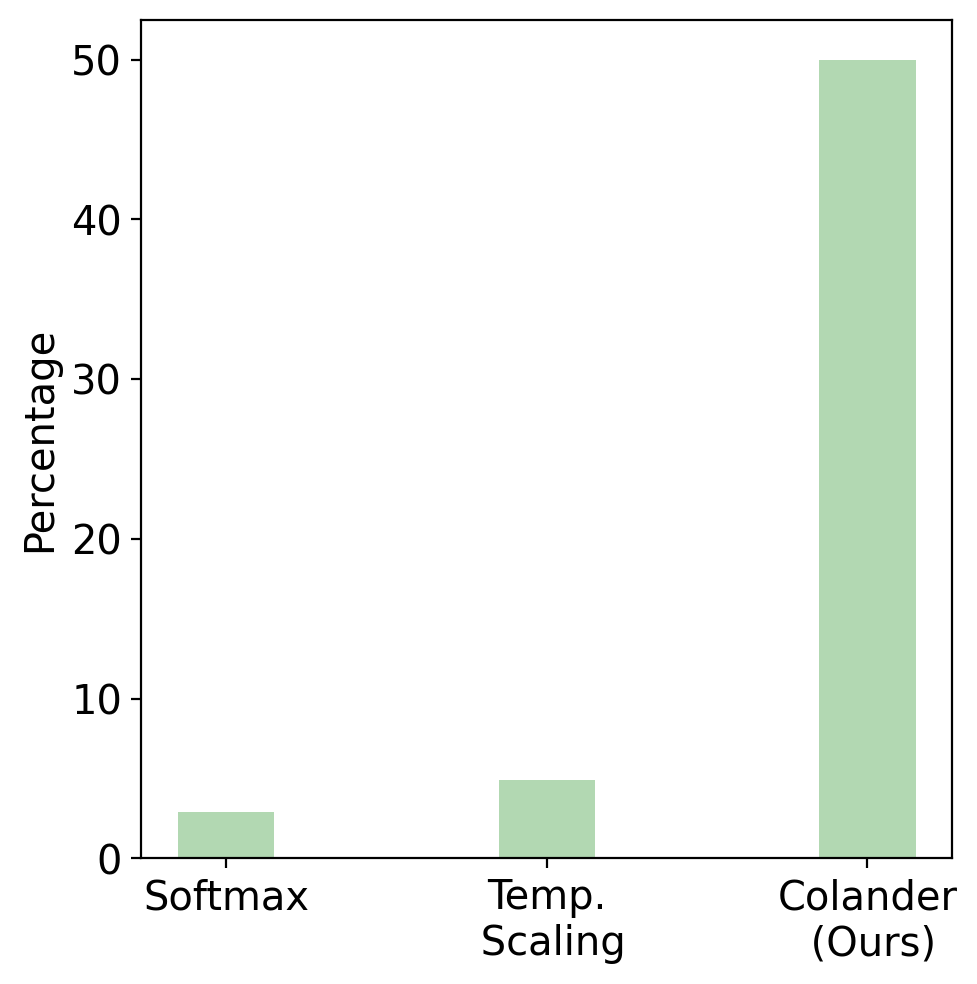}}

    \subfigure[ Auto-labeling error\label{fig:error}]{\includegraphics[width=0.195\linewidth]{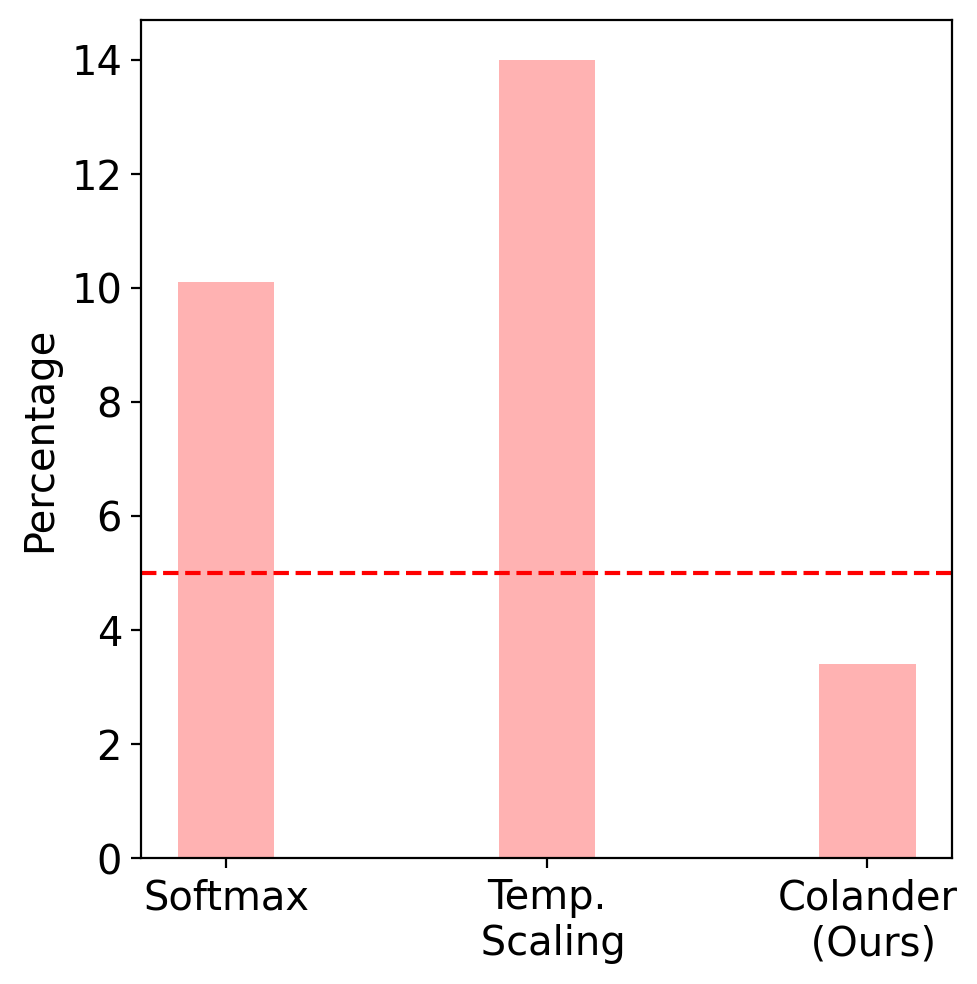}}
    
  }
  
  \caption{  {
  Scores distributions (Kernel Density Estimates) of a CNN model trained on CIFAR-10 data.  (a) softmax scores of vanilla training procedure (SGD)  (b) scores after post-hoc calibration using temperature scaling and (c) scores from our \ourmethod\, procedure applied on the same model. For training the CNN model we use 4000 points drawn randomly, and the number of validation points is 1000 (of which 500 are used for Temp. Scaling and  \ourmethod\,). The test accuracy of the model is 55\%. Figures (d) and (e) show the coverage and auto-labeling error of these methods. The dotted-red line corresponds to a 5\% error threshold.
    }}
  \label{fig:scores_dist}
\end{figure*}

The confidence function is critical to the TBAL workflow (Figure \ref{fig:auto-labeling-io}).
Existing TBAL systems rely on commonly used functions like softmax outputs from neural network models ~\citep{qiu2020min-cost-al, vishwakarma2023promises}. These functions \emph{are not well aligned with the objective of the auto-labeling system}. Using them results in substantially suboptimal coverage (Figure \ref{fig:default_scores_dist}). 
%
Hence, a query arises:

\begin{tcolorbox}[top=3pt,bottom=3pt,left=40pt,right=3pt, colback={gray!8!white}]
 What are the right choices of confidence functions for TBAL and how can we obtain such functions? 
\end{tcolorbox}

An ideal confidence function for auto-labeling will achieve the maximum coverage at a given auto-labeling error tolerance and thus will bring down the labeling cost significantly. 
Finding such an ideal function, however, is difficult because of the \emph{inherent tension} between accuracy and coverage. 
The models used in auto-labeling are often highly inaccurate so achieving a certain error guarantee is easier when being conservative in terms of confidence---but this reduces coverage.
Conversely, high coverage may appear to require lowering the requirements in confidence, but this may easily lead to overshooting the error bar.
This is compounded by the fact that TBAL is iterative so that even small deviations in tolerable error levels can cascade in future iterations. 

Worse yet, overconfidence may ruin any hope of balancing accuracy and coverage. Furthermore, in TBAL the models are trained on a small amount of labeled data. Hence, the models are not highly accurate, making the problem of designing functions for such models even more challenging.

Commonly used training procedures produce overconfident scores---high scores for both correct and incorrect predictions 
~\citep{szegedy2014Intriguing, nguyen2015deep, hendrycks2017OrdinalOOD, hein2018Confidence, bai2021OverConfLogisticReg}. Figure \ref{fig:default_scores_dist} shows that the softmax scores are overconfident, resulting in poor auto-labeling performance. Several methods have been introduced to overcome overconfidence, including calibration methods ~\citep{guo2017calibration}. Using them still misses out on significant performance (Figure \ref{fig:calib_scores}) since the calibration goal differs from auto-labeling. From the auto-labeling standpoint, we want minimum overlap between the correct and incorrect model prediction scores. Other solutions ~\citep{corbi2019Confidnet,moon2020CRL}  either bake the objective of separating scores into model training or use different optimization procedures ~\citep{zhu2022FMFP} that can encourage such separation. We observe that these do not help TBAL as well since, after some point, the model is correct on almost all the training points, making it hard to train it to discriminate between its own correct and incorrect predictions.



We address these challenges by proposing a framework to learn the right confidence functions for TBAL. In particular, we express the auto-labeling objective as an optimization problem over the space of confidence functions and the thresholds. Our framework subsumes existing methods ---they become points in the space of solutions. 
We introduce \ourmethod\, (Confidence functions for Efficient and Reliable Auto-labeling) based on a practical surrogate to the framework that can be used to learn optimal confidence functions for auto-labeling. Using these learned functions in the TBAL can achieve up to 60\% improvements in coverage versus baselines like softmax, temperature scaling ~\citep{guo2017calibration}, CRL ~\citep{moon2020CRL} and FMFP ~\citep{zhu2022FMFP}.

\vspace{10pt}
We summarize our contributions as follows,
\begin{enumerate}[topsep=2pt,itemsep=0.5ex,partopsep=1ex,parsep=1ex,leftmargin=18pt]

 \item We propose a principled framework to study the choices of confidence functions suitable for auto-labeling and provide a practical method (\ourmethod) to learn confidence functions for efficient and reliable auto-labeling.
 
\item We systematically study commonly used choices of scoring functions and calibration methods and demonstrate that they lead to poor auto-labeling performance. 
\item Through extensive empirical evaluation on real data, we show that using the confidence scores obtained using our procedure boosts auto-labeling performance significantly in comparison to common choices of confidence functions and calibration methods.
\end{enumerate}

\section{Background and Motivation}

\label{sec:TBALdetails}
We begin with setting up some useful notation. 

\textbf{Notation.} Let $[m] := \{1, 2, \ldots , m\}$ for any natural number $m$. Let $X_{u}$ be a set of unlabeled points drawn from some instance space $\cX$. Let $\cY = \{1,\ldots, k\}$ be the label space and let there be an unknown groundtruth labeling function $\ff^*:\mathcal{X} \to \mathcal{Y}$. 
Let $\cO$ be a \emph{noiseless} oracle that provides the true label for any point $\vx \in \cX$. Denote the model (hypothesis) class of classifiers by $\cH$, where each $\fh \in \cH $ is a function $\fh:\cX \to \cY$. Each classifier $h$ also has an associated \emph{confidence function} $\fg: \mathcal{X} \to \Delta^k$ that quantifies the confidence of the prediction by model $\fh \in \cH$ on any data point $\vx \in \mathcal{X}$. Here, $\Delta^k$ is a $(k-1)$-dimensional probability simplex. Let $\bv[i]$ denote the $i^\text{th}$ component for any vector $\bv \in \R^d$. For any point $\vx \in \cX$ the prediction is $\hat{y} \ldef \fh(\vx)$ and the associated confidence is $\fg(\vx)[\hat{y}]$. The vector $\vt$ denotes scores over $k$-classes, and $\vt[y]$ denotes its $y^\text{th}$ entry, i.e., score for class $y$. Please see Table \ref{table:glossary} for a summary of the notation.

\subsection{Threshold-based Auto-labeling}
\label{subsec:background}
Threshold-based auto-labeling (TBAL) (Figure \ref{fig:auto-labeling-io}) seeks to obtain labeled datasets while reducing the labeling burden on humans. The input is a pool of unlabeled data $X_u$. It outputs, for each $\vx \in X_{u}$, label $\tilde{y} \in \cY $. The output label could be either $y$, from the oracle (representing a human-obtained label), or $\hat{y}$, from the model. Let $N_u$ be the number of unlabeled points, $A\subseteq [N_u]$ the set of indices of auto-labeled points, and $X_{u}(A)$ be these points. Let $N_a$ denote the size of the auto-labeled set $A$. The \emph{auto-labeling error} denoted by $\haterr(X_{u}(A))$ and the \emph{coverage} denoted by  $\hatcov(X_{u}(A))$ of the TBAL are defined as follows:

\vspace{-20pt}
\begin{multicols}{2}
\begin{equation}
  \haterr(X_{u}(A)) \ldef \frac{1}{N_a}\sum_{i \in A}  \mathds{1}(\tilde{y}_i \neq \ff^*(\vx_i)),
  \label{eq:auto-err}
\end{equation} \break
\begin{equation}
   \hatcov(X_{u}(A)) \ldef \frac{|A|}{N_u} = \frac{N_a}{N_u},
   \label{eq:auto-cov}
\end{equation}
\end{multicols}
\vspace{-5pt}

The goal of an auto-labeling algorithm is to label the dataset so that $\haterr(X_{u}(A)) \le \epsilon_a$ while maximizing coverage $\hatcov(X_{u}(A))$ for any given $\epsilon_a \in [0, 1]$. As depicted in Figure \ref{fig:auto-labeling-io} the TBAL algorithm proceeds iteratively. In each iteration, it queries labels for a subset of unlabeled points from the oracle. It trains a classifier from the model class $\mathcal{H}$ on the oracle-labeled data acquired till that iteration. It then uses the model's confidence scores on the validation data to identify the region in the instance space, where the current classifier is confidently accurate and automatically labels the points in this region.

\subsection{Problems with confidence functions in TBAL}
\label{subsec:motivation}
The success of TBAL hinges significantly on the ability of the confidence scores of the classifier to distinguish between correct and incorrect labels. Prior works on TBAL ~\citep{vishwakarma2023promises, qiu2020min-cost-al} train the model with Stochastic Gradient Descent (SGD) and use the softmax output of the model as confidence scores which are known to be overconfident~\citep{ nguyen2015deep}. 
A natural choice to mitigate this problem is to use post-hoc calibration techniques, e.g., temperature scaling ~\citep{guo2017calibration}. 
We evaluate these choices by running TBAL for a single round on the CIFAR-10 ~\citep{krizhevsky2009CIFAR-10} dataset with a SimpleCNN model with 5.8M parameters ~\citep{mednet} with error threshold $5\%$. See Appendix \ref{subsec:motivation_exp_details} for more details.

In Figures \ref{fig:coverage} and \ref{fig:error} we observe that using softmax scores from the classifier only produces $2.9\%$ coverage while the error threshold is violated with $10\%$ error. Using temperature scaling only increases the coverage marginally to $4.9\%$ and still violates the threshold with error $14\%$. Looking closer at the scores for correct versus incorrect examples on validation data, we observe a large overlap for softmax (Figure \ref{fig:default_scores_dist}) and a marginal shift with considerable overlap for temperature scaling (Figure \ref{fig:calib_scores}). To overcome this challenge, we propose a novel framework (Section~\ref{section:OurMethod}) to learn such confidence functions in a principled way. Our method in this example can achieve $50\%$ coverage with an error of $3.4\%$ within the desired threshold (Figure \ref{fig:colander_scores}).

\section{Proposed Method (\ourmethod)}

\begin{figure*}[t]
    \centering
    
\includegraphics[width=0.96\linewidth]{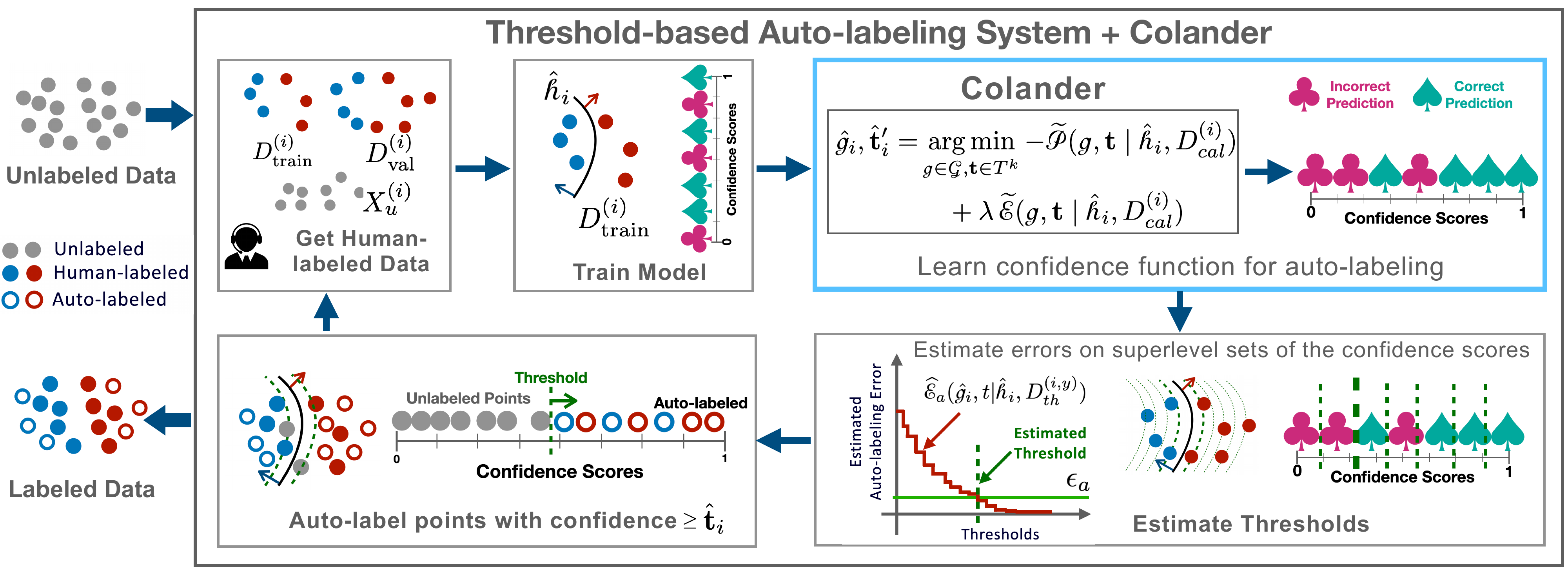}
    \vspace{2pt}
    \caption{Threshold-based Auto-labeling with \ourmethod. Similar to the existing TBAL  (Figure \ref{fig:auto-labeling-io}) it takes unlabeled data as input, selects a small subset of data points, and obtains human labels for them to create $\Dtraini$ and $\Dvali$ for the $i$th iteration. Then it trains model $\hat{\fh}_i$ on $\Dtraini$. In contrast to the standard TBAL procedure, here we randomly split $\Dvali$ into two parts $\Dcali$ and $\Dthi$. Then  \ourmethod\ kicks in, it takes $\hat{\fh}_i$ and $\Dcali$ as input and learns coverage maximizing confidence function  $\hat{\fg}_i$ for $\hat{\fh}_i$. Then using $\Dthi$ and $\hat{\fg}_i$ auto-labeling thresholds $\hat{\vt}_i$ are determined to ensure the auto-labeled data as error at most $\epsilon_a$. After obtaining the thresholds the rest of the steps are the same as the standard  TBAL. The whole workflow runs in a loop until all the data is labeled or some other stopping criteria are achieved.  }
    \label{fig:tbal-colander-workflow}
\end{figure*}

\label{section:OurMethod}
The observations in Figure \ref{fig:default_scores_dist} and \ref{fig:calib_scores} suggest that arbitrary choices of confidence functions can leave significant coverage on the table. To find a better choice of confidence function in a principled manner, we develop a framework based on auto-labeling objectives---maximizing coverage while having bounded auto-labeling error. We instantiate it by using empirical estimates and easy-to-optimize surrogates. We use the overall TBAL workflow from \citet{vishwakarma2023promises} and introduce our method to replace the confidence (scoring) function after training the classifier.


\subsection{Auto-labeling optimization framework} 
In any iteration of TBAL, we have a model $\fh$ trained on a subset of data labeled by the oracle. This model may not be highly accurate. However, it could be accurate in some regions of the instance space, and with the help of a confidence function $\fg$, we want to identify the points where the model is correct and auto-label them. As we saw earlier, arbitrary choices of $\fg$ perform poorly on this task. Instead of relying
on these choices, we propose a framework to find the right function from a sufficiently rich family of confidence functions that also subsumes the current choices.

\textbf{Optimal confidence function. } To find the confidence function aligned with our objective, we consider a rich enough space of the confidence functions $\cG$  and thresholds and express the auto-labeling objective as an optimization problem \eqref{eq:true_opt} over these spaces.

\begin{tcolorbox}[top=2pt,bottom=3pt,left=3pt,right=3pt, colback={gray!6!white}]
\begin{equation}
  \argmax_{\fg \in \cG, \vt \in T^k} \quad  \cov(\fg,\vt \given \fh) 
\; \text{ s.t.} \; \; \err(\fg,\vt \given \fh) \le \epsilon_a .
 \tag{P1}
 \label{eq:true_opt}
\end{equation}
\end{tcolorbox}

Here $T$ is the set of confidence thresholds and $\cG:\cX \to T^k$ is the set of confidence functions and $\cov(\fg, \vt |\fh)$ and $\err(\fg,\vt \given \fh) $ are the population level coverage and auto-labeling error which are defined as follows,

\vspace{-20pt}
\begin{multicols}{2}  
\begin{equation}
\label{eq:true_cov}
\cov ( \fg,\vt \given \fh) \ldef  \P_{\vx} \big(\fg(\vx)[\hat{y}]\ge \vt[\hat{y}] \big ),  
\end{equation} \break
\begin{equation}
\label{eq:true_err}
\err(\fg, \vt \given \fh) \ldef  \P_{\vx}\big(y \neq \hat{y} \given \fg(\vx)[\hat{y}] \ge \vt[\hat{y}] \big). 
\end{equation}
\end{multicols}

The optimal $\fg^\star$ and $\vt^\star$ that achieve the maximum coverage while satisfying the auto-labeling error constraint belong to the solution(s) of the following optimization problem. 
 

\subsection{Practical method to learn confidence functions}
The above framework provides a theoretical characterization of the optimal confidence functions and thresholds for TBAL. However, it is not practical since, in practice, the data distributions and $\ff^\star$ are not known. Next, we provide a practical method based on the above framework to learn confidence functions for TBAL.



\textbf{Empirical optimization problem.}
Since we do not know the distributions of $\vx$ and $\ff^\star$, we use estimates of coverage and auto-labeling errors on a fraction of validation data to solve the optimization problem.
Let $D$ be some finite number of labeled samples, and then the empirical coverage and auto-labeling error are defined as follows, 
   
    \begin{equation}
    \label{eq:est_cov}
        \hatcov(\fg,\vt \given \fh, D) \ldef \frac{1}{|D|} \sum_{(\vx,y) \in D } \one \big( \fg(\vx)[\yhat]\ge \vt[\yhat] \big), 
        \end{equation}

        \begin{equation}
        \label{eq:est_err}
\haterr(\fg, \vt \given \fh,D ) \ldef  \frac{ \sum_{(\vx,y) \in D } \one \big( y \neq \hat{y} \land \fg(\vx)[\yhat]\ge \vt[\yhat] \big) }{\sum_{(\vx,y) \in D } \one \big( \fg(\vx)[\yhat]\ge \vt[\yhat] \big)}.
    \end{equation}

We randomly split the validation data into two parts $\Dcal$ and $\Dth$ and use $\Dcal$ to compute $\hatcov(\fg, \vt \given \fh, \Dcal)$ and  $\haterr(\fg, \vt \given \fh, \Dcal)$ for the following empirical version of the optimization problem. We now hope to solve the following optimization problem using these estimates to get $\hat{\fg}, \hat{\vt}$.
\begin{tcolorbox}[top=2pt,bottom=3pt,left=0pt,right=0pt, colback={gray!6!white}]
\begin{equation}
  \argmax_{\fg \in \cG, \vt \in T^k}  \quad \hatcov(\fg,\vt \given \fh, \Dcal ) 
\; \text{ s.t.} \; \; \haterr(\fg,\vt \given \fh, \Dcal) \le \epsilon_a.
 \tag{P2}
 \label{eq:emp_opt}
\end{equation}
\end{tcolorbox}

However, there is a caveat: the objective and constraint are based on $0\text{-}1$ variables, so it is hard to optimize for $\fg$ and $\vt$.

\textbf{Surrogate optimization problem. }
To make the above optimization \eqref{eq:emp_opt}  tractable using gradient-based methods, we introduce differentiable surrogates for the $0\text{-}1$ variables. Let $\sigma(\alpha, z) \ldef 1/ ({1 + \exp(-\alpha z)}) $ denote the sigmoid function on $\R$ with scale parameter $\alpha \in \R$. It is easy to see that, for any $\fg, y$ and $\vt$, $\fg(\vx)[y] \ge \vt[y] \iff \sigma(\alpha, \fg(\vx)[y]-\vt[y]) \ge 1/2$. Using this fact, we define the following surrogates of the auto-labeling error and coverage as follows,

{
    
    \begin{equation}
    \label{eq:surr_cov}
        \surrcov(\fg,\vt | \fh, \Dcal) \ldef \frac{1}{|\Dcal|} \sum_{(\vx,y) \in \Dcal} \sigma \big( \alpha, \fg(\vx)[\yhat]- \vt[\yhat] \big) ,
    \end{equation}
    
    }

\begin{equation}
\label{eq:surr_err}
\surrerr(\fg,\vt \given \fh, \Dcal ) \ldef  \frac{ \sum_{(\vx,y) \in \Dcal } \one \big( y \neq \yhat \big) \, \sigma \big( \alpha, \fg(\vx)[\yhat]- \vt[\yhat] \big)  }{\sum_{(\vx,y) \in \Dcal } \sigma\big( \alpha, \fg(\vx)[\yhat]- \vt[\yhat] \big) }.
\end{equation}
   
and the surrogate optimization problem as follows,
\begin{tcolorbox}[top=2pt,bottom=3pt,left=3pt,right=3pt, colback={gray!8!white}]
\begin{equation}
  \argmin_{\fg \in \cG, \vt \in T^k} \quad  -\surrcov(\fg,\vt \given \fh, \Dcal) 
+ \lambda \,\surrerr(\fg,\vt \given \fh, \Dcal) 
 \tag{P3}
 \label{eq:surr_opt}
\end{equation}
\end{tcolorbox}
Here, $\lambda \in \R^+$ is the penalty term controlling the relative importance of the auto-labeling error and coverage. It is a hyper-parameter, and we find it using our hyper-parameter searching procedure discussed in section \ref{subsec:hyp-search}. The gap between the surrogate and actual coverage, error diminishes as $\alpha \to \infty$. We discuss this in Appendix \ref{sec:method_details}.





 \setlength{\textfloatsep}{8pt}
\begin{algorithm}[h]

 \caption{Threshold-based Auto-Labeling (TBAL)}
     \vspace{3pt}

  \textbf{Input:} Unlabeled data $X_{u}$, labeled validation data $\Dval$, auto labeling error tolerance $\epsilon_a$, $N_t$ training data query budget, seed data size $n_s$, batch size for active query $n_b$, calibration data fraction $\nu$, set of confidence thresholds $T$, coverage lower bound $\rho_0$, label space $\mathcal{Y}.$ 
       \vspace{2pt}

  \textbf{Output:} Auto-labeled dataset $D_\mathrm{out}$
  
 \begin{algorithmic}[1]
     \vspace{3pt}

    \Procedure{TBAL}{$X_u, D_\mathrm{val}, \epsilon_a, N_t, n_s, n_b, \nu, \rho_0, T, \mathcal{Y}$}
        \vspace{2pt}

    \LineComment{/*** Initialization. ***/}
    \vspace{2pt}
    \State $D_\mathrm{query}^{(1)} \gets \Call{RandomQuery}{X_u, n_s}$ \Comment{ Randomly select $n_s$ points and get manual labels for them.}
        \vspace{2pt}
    \State $X_u^{(1)} \gets X_{u} \setminus \{ \vx : (\vx, y) \in D_\mathrm{query}^{(1)}\}$ \Comment{Remove the manually labeled points from the unlabeled pool.}
        \vspace{2pt}

    \State $D_\mathrm{val}^{(1)} \gets D_{\mathrm{val}};  D^{(0)}_\mathrm{train} \gets \emptyset$ \Comment{Validation data for the first round is full $\Dval$. }
    \vspace{3pt}
    \State $D_\mathrm{out} \gets D_\mathrm{query}^{(1)}; n_t^{(1)} \gets n_s ; i \gets 1$ \Comment{Include the manually labeled data in Step 2. in the output data $D_\mathrm{out}$.}
    \vspace{5pt}
       \LineComment{/*** Run the auto-labeling loop ***/ }
       \LineComment{/* Until no more unlabeled points are left or the budget for manually labeled training data is exhausted. */}
       \vspace{2pt}
    \While {$X_u^{(i)} \neq \emptyset$ and $n_t^{(i)} \le N_t$} 
    \vspace{2pt}
    \State $D^{(i)}_\mathrm{train}  \hspace{18pt} \gets D^{(i-1)}_\mathrm{train} \cup D^{(i)}_\mathrm{query}$ \Comment{Include the manually labeled points in the training data.}
    \vspace{2pt}
    \State $\hat{\fh}_i \hspace{35pt} \gets \Call{TrainModel}{\mathcal{H}, D_\mathrm{train}^{(i)}}$ \Comment{Train a classification model.}
\vspace{2pt}

    \State $\Dcali, \Dthi  \hspace{3pt} \gets \Call{RandomSplit}{D^{(i)}_\mathrm{val}, \nu}$
\vspace{2pt} \Comment{Randomly split the current validation data into two parts.}
\vspace{5pt}
\LineComment{/*** \ourmethod\,  block, to learn the new confidence function $\hat{\fg}_i$***/}
\vspace{5pt}
    \State $
       {\color{blue} \hat{\fg}_i, \hat{\vt}'_i  \hspace{10pt}\gets \argmin_{\fg \in \cG, \vt \in T^k}  -\surrcov(\fg,\vt \given \hat{\fh}_i, \Dcali)
                                     + \lambda \,\surrerr(\fg,\vt \given \hat{\fh}_i, \Dcali)}
    $ \Comment{ \ourmethod\, procedure.}

    \vspace{2pt}
    

    \LineComment{/*** Estimate auto-labeling thresholds using  $\hat{\fg}_i$ and $\Dthi$. See Algorithm \ref{alg:tbal-conf-gamma-estimate}. ***/}
    \vspace{3pt}
    \State $\hat{\vt}_i  \hspace{18pt} \gets \Call{EstThreshold}{\hat{\fg}_i,\hat{\fh}_i, \Dthi, \epsilon_a, \rho_0, T, \mathcal{Y}}$  
    \vspace{3pt}

    
\LineComment{ /*** Auto-label the points having scores above the thresholds. ***/}
\vspace{2pt}

\State $\widetilde{D}_u^{(i)}  \hspace{10pt}\gets \{(\vx, \hat{\fh}_i(\vx)) : \vx \in X^{(i)}_u\}$ 
\vspace{2pt}
    \State $D_\mathrm{auto}^{(i)}  \hspace{3pt} \gets \{  (\bx,\hat{y}) \in \tilde{D}_u^{(i)} : \hat{\fg}_i(\bx)[\,\hat{y}]\, \ge \hat{\vt}_i[\, \hat{y} ]\, \}$
    \vspace{2pt}
    \State $X_u^{(i)}  \hspace{8pt}\gets X_u^{(i)} \setminus \{ \vx : (\vx, \hat{y}) \in D_\mathrm{auto}^{(i)}\}$ \Comment{Remove auto-labeled points from the unlabeled pool.}
    \vspace{2pt}
    \State $\widetilde{D}_{\mathrm{val}}^{(i)}  \hspace{8pt} \gets \{(\vx, \hat{\fh}_i(\vx)) : (\vx,y) \in \Dvali \}$ 

    \vspace{2pt}
    \State $D_\mathrm{val}^{(i+1)} \gets \{(\bx,\hat{y}) \in \tilde{D}_\mathrm{val}^{(i)} : \hat{\fg}_i(\bx)[\hat{y}] < \hat{\vt}_i[\hat{y}]  \} $ \Comment{Remove validation points in the auto-labeling region.}
    \vspace{3pt}
     \LineComment{/*** Get the next batch of manually labeled data using an active querying strategy. ***/}
     \vspace{2pt}
    \State $D_\mathrm{query}^{(i+1)} \gets \Call{ActiveQuery}{\hat{\fh}_i, X_u^{(i)},n_b}$
        \vspace{2pt}
    \State $X_u^{(i+1)} \gets X_u^{(i)} \setminus \{ \vx : (\vx, y) \in D_\mathrm{query}^{(i + 1)}\}$ \Comment{Remove manually labeled data from the unlabeled pool.}
    \vspace{2pt}
    \State $D_\mathrm{out}  \hspace{7pt}\gets D_\mathrm{out} \cup D_\mathrm{auto}^{(i)} \cup D_\mathrm{query}^{(i+1)}$ \Comment{Add the auto-labeled and manually labeled points in the output data.}
    \vspace{2pt}
    \State $ n_t^{(i+1)} \hspace{2pt} \gets n_t^{(i)} + n_b $
    \State $i \hspace{6pt} \gets i +1$ 
    \EndWhile
    \State \textbf{return } $D_\mathrm{out}$
    
    \EndProcedure
 \end{algorithmic}
 \label{alg:main_algo}
 \end{algorithm}

\begin{algorithm}[h]
\caption{Estimate Auto-Labeling Threshold}
    \textbf{Input:} Confidence function $\hat{\fg}_i$, classifier $\hat{\fh}_i$, Part of validation data $\Dthi$ for threshold estimation, auto labeling error tolerance $\epsilon_a$, set of confidence thresholds $T$, coverage lower bound $\rho_0$, label space $\mathcal{Y}$.\\
    \textbf{Output:}   Auto-labeling thresholds $\hat{\vt}_{i}$, where $\hat{\vt}_i[y]$ is the threshold for class $y$.
\begin{algorithmic}[1]
    \Procedure{EstThreshold}{$\hat{\fg}_i,\hat{\fh}_i, \Dthi, \epsilon_a, \rho_0, T, \mathcal{Y}$}    
  \vspace{2pt}
  \LineComment{/*** Estimate thresholds for each class. ***/}
  \vspace{2pt}
    \For{ $ y \in \cY $}
        \vspace{2pt}
        \State $\Dthiy \gets \{ (\vx',y') \in \Dthi : y'=y \}$ \Comment{Group points class-wise.}
        \vspace{2pt}
        \LineComment{ /*** Only evaluate thresholds with est. coverage at least $\rho_0$. ***/}
        \vspace{1pt}

        \State $T'_y \hspace{15pt} \gets \{  t \in T : \hatcov \big(\hat{\fg}_i ,t\given \hat{\fh}_i, \Dthiy \big) \ge \rho_0 \} \cup \{ \infty \} $ 
        \vspace{2pt}
        \LineComment{/*** Estimate auto-labeling error at each threshold. Pick the smallest threshold with the sum of estimated error and $C_1$ times the standard deviation is below $\epsilon_a$. $C_1$ is set to 0.25 here. ***/}
        \vspace{2pt}
        \State $\hat{\vt}_i[y] \hspace{6pt} \gets \min \{ t \in T'_y : \haterr_a(\hat{\fg}_i, t |\hat{\fh}_i, \Dthiy)
                                 + C_1\hat{\sigma}(\hat{\fh}_i,t, \Dthiy)  \le \epsilon_a \}$
        \vspace{3pt}
    \EndFor
    \State \textbf{return } $\hat{\vt}_i$
    
    \EndProcedure
    \end{algorithmic}
    \label{alg:tbal-conf-gamma-estimate}
\end{algorithm}

\begin{wrapfigure}{r}{0.45\linewidth}
\vspace{-10pt}
    \begin{center}
        \includegraphics[width=0.9\linewidth]{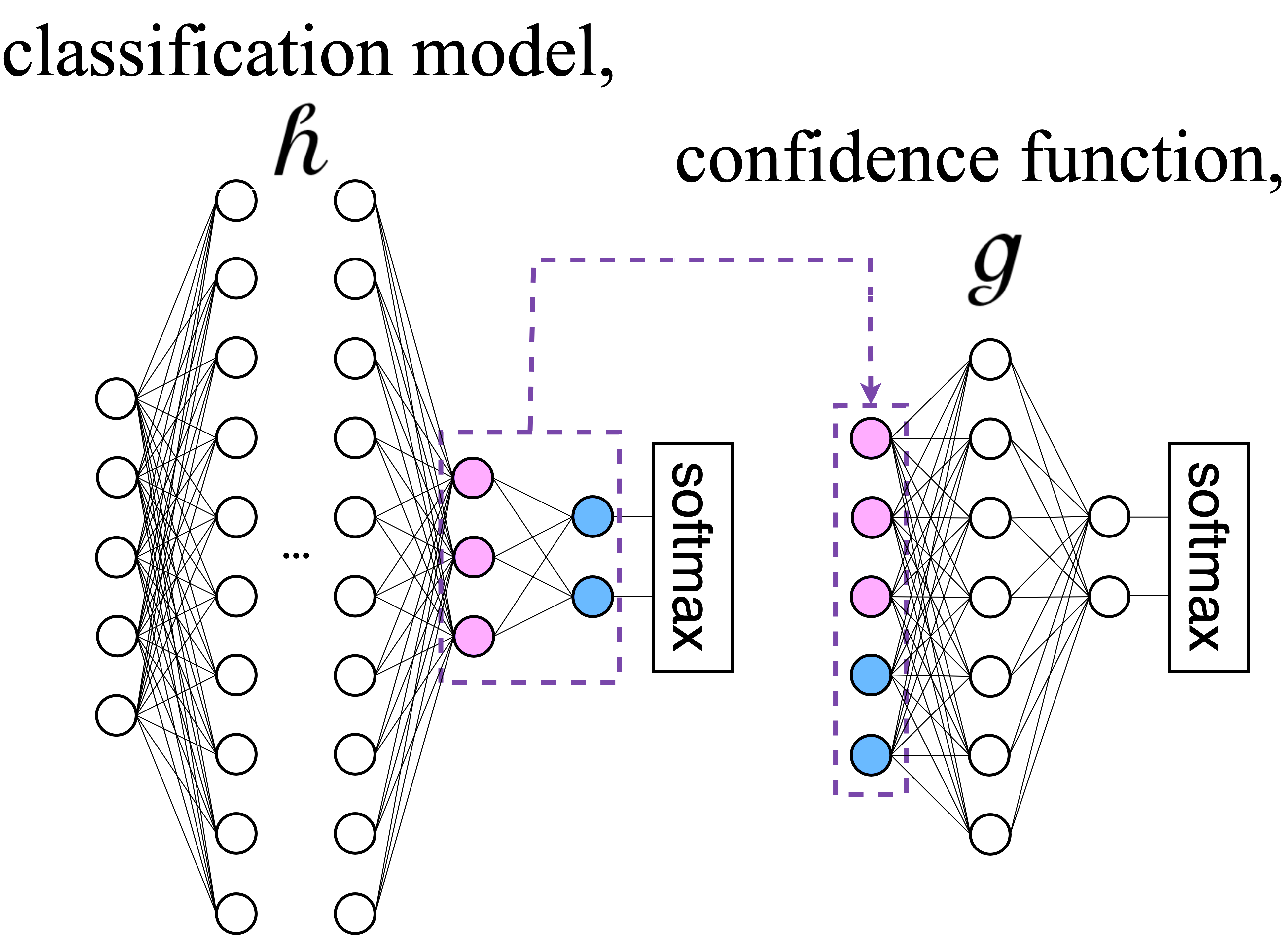}
    \end{center}
    \caption{Our choice of $\fg$ function.}
    \label{fig:illu_our}
    \vspace{-8pt}
\end{wrapfigure}

\textbf{Choice of $\cG$.}
Our framework is flexible with the function class $\cG$ choice. In this work, we use deep neural networks (DNNs) with at least two layers on model class $\cH$. Since DNNs also learn powerful representations during training, we use the last two layers of representations as input for the functions in $\cG$ (Figure \ref{fig:illu_our}). Let $\bz^{(1)}{(\bx;\fh)} \in \R^{k}$ and $\bz^{(2)}(\bx;h) \in \R^{d_2}$ be the outputs(logits) of the last and the second-last layer of the net $h$ for input $\vx$ and let $\bz(\bx;h) \ldef [\bz^{(1)}(\bx;\fh), \bz^{(2)}(\bx;\fh)]$ denote the concatenation of the two representations. We propose to use two-layer neural networks $\cG_{nn_2} : \R^{k+d_2} \mapsto \Delta^{k}$ for $\cG$. A net $\fg \in \cG_{nn_2}$ takes the last two layer's representations from $h$ and outputs confidence scores over $k$ classes. Given $\fh$, the $\fg$ is defined as follows,

\begin{equation}
    \fg(\bx) \ldef \texttt{softmax}\big( \mathbf{W}_2\texttt{tanh} (\mathbf{W}_1 \bz(\bx;\fh)) \big). 
\end{equation}

Here $\mathbf{W}_1 \in \R^{(k+d_2) \times 2(k+d_2)}$ and $\R^{ 2(k+d_2)\times k}$ are the learnable weight matrices and for any $\vv \in \R^d$, the $\texttt{softmax}(\vv)[i] \ldef \exp(\vv[i])/ (\sum_j \exp(\vv[j]))$ and $\texttt{tanh}(\vv)[i] \ldef (\exp(2\vv[i])-1)/(\exp(2\vv[i])+1)$.






\textbf{Solving the surrogate optimization. }
The optimization problem \eqref{eq:surr_opt} is non-convex even for a simple class of $\fg$ (such as linear). Nevertheless, it is differentiable and we apply gradient-based methods, which have been highly effective in minimizing non-convex losses in deep learning. We solve for $\fg$ and $\vt$ simultaneously using the Adam optimizer ~\citep{kingma2014adam}. The details of the training hyperparameters are deferred to the Appendix \ref{subsec:hyperparameters}.

\vspace{-5pt}
\subsection{TBAL procedure with \ourmethod}
 We plugin our method \ourmethod\ to learn confidence functions in the workflow of TBAL (Algorithm \ref{alg:main_algo}). The workflow is also illustrated in Figure \ref{fig:tbal-colander-workflow}.  The steps in the updated workflow are the same as the standard TBAL (Figure \ref{fig:auto-labeling-io}), except for the introduction of \ourmethod\ after the model training step to learn a new confidence function $\hat{\fg}_i$ using part of the validation data ($\Dcali$) and then the threshold estimation procedure (Algorithm \ref{alg:tbal-conf-gamma-estimate}) finds auto-labeling thresholds $\hat{\vt}_i$ on the scores computed using $\hat{\fg}_i$ on the other part of the validation data called $\Dthi$. While we get thresholds as output from \ourmethod, it is important to estimate them again from the held-out data $\Dthi$ to ensure the auto-labeling error constraint is not violated. In Algorithm \ref{alg:main_algo} the procedure $\Call{RandomQuery}{X_u, n_s}$ selects $n_s$ points randomly from $X_u$ and obtains human labels for them to create $D_\mathrm{train}^{(1)}$. The procedure $\Call{RandomSplit}{\Dvali,\nu}$ randomly splits $\Dvali$, the validation data in $i^{\mathrm{th}}$ iteration to $\Dcali$ and $\Dthi$ with $\Dcali$ having $\nu$ fraction of points from $\Dvali$. $\Dcali$ is used for learning the post-hoc confidence function and $\Dthi$ is used for estimating auto-labeling thresholds in Algorithm \ref{alg:tbal-conf-gamma-estimate}. The procedure,
 $\Call{TrainModel}{\cH,D^{(i)}_\mathrm{train}}$ trains a model from model class $\cH$ on the training data $D^{(i)}_\mathrm{train}$. Any training procedure can be used here, in this work we use methods listed in Section \ref{subsubsec:train-time-methods} for model training. Lastly, $\Call{ActiveQuery}{\hat{\fh}_i, X_u^{(i)},n_b}$, selects $n_b$ points from the remaining unlabeled pool using the same active learning strategy used in a prior work ~\citep{vishwakarma2023promises}. We defer the details to the Appendix \ref{sec:method_details}.

\section{Empirical Evaluation}
\label{sec:exp-main}

As we observed in Section \ref{subsec:motivation}, ad-hoc choices of confidence functions can lead to poor auto-labeling performance. Motivated by these shortcomings, we designed a method to learn confidence functions that are well-aligned with the auto-labeling objective. In this section, we verify the following claims through extensive empirical evaluation,  

\textbf{C1.} \ourmethod\, learns better confidence functions for auto-labeling compared to standard training and common post-hoc methods that mitigate the overconfidence problem. Using it in TBAL can boost the coverage significantly while keeping the auto-labeling error low.

    \textbf{C2.} \ourmethod\, is not dependent on any particular train-time method and thus should improve the performance over using any train-time method alone.  
\subsection{Baselines} \label{sec:baselines}
We compare several train-time and post-hoc methods that improve confidence functions from calibration and ordinal ranking perspectives. Detailed descriptions of these methods are deferred to Appendix \ref{sec:train-time-post-hoc}.

\subsubsection{Train-time methods}
\label{subsubsec:train-time-methods}
We use the following methods for training the model $\hat{\fh}$.
\begin{enumerate}[topsep=0pt,itemsep=-0.5ex,partopsep=1ex,parsep=1ex,leftmargin=24pt]
\item \textit{Vanilla} neural network trained under cross-entropy loss using stochastic gradient descent (SGD) ~\citep{amari1993backpropagation, Bottou2012, guo2017calibration}.


\item \textit{Squentropy} ~\citep{hui2023Squentropy} adds the average square loss over the incorrect classes to cross-entropy loss to improve the calibration and accuracy of the model. 

\item \textit{Correctness Ranking Loss (CRL)} ~\citep{moon2020CRL} aligns the confidence scores of the model with the ordinal rankings criterion via regularization. 

\item \textit{FMFP} ~\citep{zhu2022FMFP} aligns confidence scores with the ordinal rankings criterion by using Sharpness Aware Minimization (SAM) \citep{foret2021sharpnessaware} in lieu of SGD. 

\end{enumerate}

\begin{table*}[t]
\begin{center}
\begin{small}
\resizebox{\textwidth}{!}{
\begin{tabular}{@{}lllllllllll@{}}
\toprule
\textbf{Dataset} & \textbf{Model $\fh$} & \textbf{$N$} & $N_u$ &\textbf{$K$} & \textbf{$N_t$} & \textbf{$N_v$} & \textbf{$N_{\mathrm{hyp}}$} & \textbf{Modality} & \textbf{Preprocess} & \textbf{Dimension}        \\ \midrule
MNIST            & LeNet-5              & 70k  & 60k     & 10           & 500            & 500            & 500                           & Image             & None                   & 1 $\times$ 28 $\times$ 28 \\
CIFAR-10         & CNN  & 50k   & 40k    & 10           & 10k         & 8k          & 2k                         & Image             & None                   & 3 $\times$ 32 $\times$ 32 \\
Tiny-Imagenet    & MLP   & 110k  & 90k    & 200          & 10k         & 8k         & 2k                         & Image             & CLIP                   & 512                       \\
20 Newsgroup     & MLP   & 11.3k    & 9k   & 20           & 2k          & 1.6k          & 600                           & Text              & FlagEmb.          & 1,024                        \\ \bottomrule
\end{tabular}
}
\end{small}
\end{center}
\caption{Details of the dataset and model we used to evaluate the performance of our method and other calibration methods. For the Tiny-Imagenet and 20 Newsgroup datasets, we use CLIP and FlagEmbedding, respectively, to obtain the embeddings of these datasets and conduct auto-labeling on the embedding space. For Tiny-Imagenet, we use a 3-layer perceptron with 1,000, 500, 300 neurons on each layer as model $\fh$; for 20 Newsgroup, we use a 3-layer perceptron with 1,000, 500, 30 neurons on each layer as model $\fh$.}
\label{tab:settings_details}
\end{table*}

    \begin{table*}[t]
    \centering
    \fontsize{9}{11}\selectfont
    \resizebox{\textwidth}{!}{
    \begin{tabular}{llcccccccc}
    \toprule \noalign{\vskip1pt}
    \multicolumn{1}{c}{\multirow{2}{*}{\textbf{Train-time}}} & \multicolumn{1}{c}{\multirow{2}{*}{\textbf{Post-hoc}}}  & \multicolumn{2}{c}{\textbf{MNIST}} & \multicolumn{2}{c}{\textbf{CIFAR-10}} & \multicolumn{2}{c}{\textbf{20 Newsgroups}} & \multicolumn{2}{c}{\textbf{Tiny-ImageNet}} \\ \noalign{\vskip1pt} \cline{3-10} \noalign{\vskip1pt} 
    \multicolumn{1}{c}{}                      & \multicolumn{1}{c}{}  &  \multicolumn{1}{c}{\cellcolor{red!20}\textbf{Err ($\downarrow$)}} & \multicolumn{1}{c}{\textbf{Cov ($\uparrow$)}} &  \multicolumn{1}{c}{\cellcolor{red!20}\textbf{Err ($\downarrow$)}} & \multicolumn{1}{c}{\textbf{Cov ($\uparrow$)}} &  \multicolumn{1}{c}{\cellcolor{red!20}\textbf{Err ($\downarrow$)}} & \multicolumn{1}{c}{\textbf{Cov ($\uparrow$)}} &  \multicolumn{1}{c}{\cellcolor{red!20}\textbf{Err ($\downarrow$)}} & \multicolumn{1}{c}{\textbf{Cov ($\uparrow$)}} \\ \noalign{\vskip1pt} \toprule \noalign{\vskip4pt} 
    \multirow{6}{*}{Vanilla}                     & Softmax &  \cellcolor{red!10}\textbf{4.1}\scalebox{0.6}{\ensuremath{\bm{\pm}}}{\fontsize{7}{11}\selectfont\textbf{0.7} } &  85.0\scalebox{0.6}{\ensuremath{\pm}}{\fontsize{7}{11}\selectfont2.5 } &  \cellcolor{red!10}4.8\scalebox{0.6}{\ensuremath{\pm}}{\fontsize{7}{11}\selectfont0.2 } &  14.0\scalebox{0.6}{\ensuremath{\pm}}{\fontsize{7}{11}\selectfont2.1 } &  \cellcolor{red!10}6.0\scalebox{0.6}{\ensuremath{\pm}}{\fontsize{7}{11}\selectfont0.6 } &  48.2\scalebox{0.6}{\ensuremath{\pm}}{\fontsize{7}{11}\selectfont1.6 } &  \cellcolor{red!10}11.1\scalebox{0.6}{\ensuremath{\pm}}{\fontsize{7}{11}\selectfont0.3 } &  32.6\scalebox{0.6}{\ensuremath{\pm}}{\fontsize{7}{11}\selectfont0.5 }\\\noalign{\vskip1pt}\hhline{~---------}\noalign{\vskip1pt}                                 &  TS &  \cellcolor{red!10}7.8\scalebox{0.6}{\ensuremath{\pm}}{\fontsize{7}{11}\selectfont0.6 } &  94.2\scalebox{0.6}{\ensuremath{\pm}}{\fontsize{7}{11}\selectfont0.5 } &  \cellcolor{red!10}7.3\scalebox{0.6}{\ensuremath{\pm}}{\fontsize{7}{11}\selectfont0.3 } &  23.2\scalebox{0.6}{\ensuremath{\pm}}{\fontsize{7}{11}\selectfont0.7 } &  \cellcolor{red!10}9.7\scalebox{0.6}{\ensuremath{\pm}}{\fontsize{7}{11}\selectfont0.6 } &  60.7\scalebox{0.6}{\ensuremath{\pm}}{\fontsize{7}{11}\selectfont2.3 } &  \cellcolor{red!10}16.3\scalebox{0.6}{\ensuremath{\pm}}{\fontsize{7}{11}\selectfont0.5 } &  37.4\scalebox{0.6}{\ensuremath{\pm}}{\fontsize{7}{11}\selectfont1.5 }\\\noalign{\vskip1pt}\hhline{~---------}\noalign{\vskip1pt}                                 &  Dirichlet &  \cellcolor{red!10}7.9\scalebox{0.6}{\ensuremath{\pm}}{\fontsize{7}{11}\selectfont0.7 } &  93.2\scalebox{0.6}{\ensuremath{\pm}}{\fontsize{7}{11}\selectfont2.2 } &  \cellcolor{red!10}7.7\scalebox{0.6}{\ensuremath{\pm}}{\fontsize{7}{11}\selectfont0.5 } &  22.4\scalebox{0.6}{\ensuremath{\pm}}{\fontsize{7}{11}\selectfont1.2 } &  \cellcolor{red!10}9.4\scalebox{0.6}{\ensuremath{\pm}}{\fontsize{7}{11}\selectfont0.9 } &  59.4\scalebox{0.6}{\ensuremath{\pm}}{\fontsize{7}{11}\selectfont1.8 } &  \cellcolor{red!10}17.1\scalebox{0.6}{\ensuremath{\pm}}{\fontsize{7}{11}\selectfont0.4 } &  33.3\scalebox{0.6}{\ensuremath{\pm}}{\fontsize{7}{11}\selectfont2.0 }\\\noalign{\vskip1pt}\hhline{~---------}\noalign{\vskip1pt}                                 &  SB &  \cellcolor{red!10}6.7\scalebox{0.6}{\ensuremath{\pm}}{\fontsize{7}{11}\selectfont0.5 } &  92.6\scalebox{0.6}{\ensuremath{\pm}}{\fontsize{7}{11}\selectfont1.5 } &  \cellcolor{red!10}6.1\scalebox{0.6}{\ensuremath{\pm}}{\fontsize{7}{11}\selectfont0.4 } &  18.6\scalebox{0.6}{\ensuremath{\pm}}{\fontsize{7}{11}\selectfont1.1 } &  \cellcolor{red!10}8.1\scalebox{0.6}{\ensuremath{\pm}}{\fontsize{7}{11}\selectfont0.6 } &  58.1\scalebox{0.6}{\ensuremath{\pm}}{\fontsize{7}{11}\selectfont1.8 } &  \cellcolor{red!10}15.7\scalebox{0.6}{\ensuremath{\pm}}{\fontsize{7}{11}\selectfont0.6 } &  35.4\scalebox{0.6}{\ensuremath{\pm}}{\fontsize{7}{11}\selectfont1.2 }\\\noalign{\vskip1pt}\hhline{~---------}\noalign{\vskip1pt}                                 &  Top-HB &  \cellcolor{red!10}7.4\scalebox{0.6}{\ensuremath{\pm}}{\fontsize{7}{11}\selectfont1.4 } &  93.1\scalebox{0.6}{\ensuremath{\pm}}{\fontsize{7}{11}\selectfont3.6 } &  \cellcolor{red!10}6.0\scalebox{0.6}{\ensuremath{\pm}}{\fontsize{7}{11}\selectfont0.7 } &  15.6\scalebox{0.6}{\ensuremath{\pm}}{\fontsize{7}{11}\selectfont1.9 } &  \cellcolor{red!10}9.2\scalebox{0.6}{\ensuremath{\pm}}{\fontsize{7}{11}\selectfont1.0 } &  59.0\scalebox{0.6}{\ensuremath{\pm}}{\fontsize{7}{11}\selectfont2.0 } &  \cellcolor{red!10}16.6\scalebox{0.6}{\ensuremath{\pm}}{\fontsize{7}{11}\selectfont0.5 } &  37.6\scalebox{0.6}{\ensuremath{\pm}}{\fontsize{7}{11}\selectfont2.2 }\\\noalign{\vskip1pt}\hhline{~---------}\noalign{\vskip1pt}                                 &  \textbf{Ours} &  \cellcolor{red!10}4.2\scalebox{0.6}{\ensuremath{\pm}}{\fontsize{7}{11}\selectfont1.5 } &  \textbf{95.6}\scalebox{0.6}{\ensuremath{\bm{\pm}}}{\fontsize{7}{11}\selectfont\textbf{1.4} } &  \cellcolor{red!10}\textbf{3.0}\scalebox{0.6}{\ensuremath{\bm{\pm}}}{\fontsize{7}{11}\selectfont\textbf{0.2} } &  \textbf{78.5}\scalebox{0.6}{\ensuremath{\bm{\pm}}}{\fontsize{7}{11}\selectfont\textbf{0.2} } &  \cellcolor{red!10}\textbf{2.5}\scalebox{0.6}{\ensuremath{\bm{\pm}}}{\fontsize{7}{11}\selectfont\textbf{1.1} } &  \textbf{80.6}\scalebox{0.6}{\ensuremath{\bm{\pm}}}{\fontsize{7}{11}\selectfont\textbf{0.7} } &  \cellcolor{red!10}\textbf{1.4}\scalebox{0.6}{\ensuremath{\bm{\pm}}}{\fontsize{7}{11}\selectfont\textbf{2.1} } &  \textbf{59.2}\scalebox{0.6}{\ensuremath{\bm{\pm}}}{\fontsize{7}{11}\selectfont\textbf{0.8} }\\\noalign{\vskip4pt} \hline \noalign{\vskip4pt}\noalign{\vskip1pt}\multirow{6}{*}{CRL}                     & Softmax &  \cellcolor{red!10}4.7\scalebox{0.6}{\ensuremath{\pm}}{\fontsize{7}{11}\selectfont0.4 } &  86.0\scalebox{0.6}{\ensuremath{\pm}}{\fontsize{7}{11}\selectfont4.5 } &  \cellcolor{red!10}5.2\scalebox{0.6}{\ensuremath{\pm}}{\fontsize{7}{11}\selectfont0.3 } &  15.9\scalebox{0.6}{\ensuremath{\pm}}{\fontsize{7}{11}\selectfont0.8 } &  \cellcolor{red!10}5.8\scalebox{0.6}{\ensuremath{\pm}}{\fontsize{7}{11}\selectfont0.5 } &  48.3\scalebox{0.6}{\ensuremath{\pm}}{\fontsize{7}{11}\selectfont0.3 } &  \cellcolor{red!10}10.4\scalebox{0.6}{\ensuremath{\pm}}{\fontsize{7}{11}\selectfont0.4 } &  32.5\scalebox{0.6}{\ensuremath{\pm}}{\fontsize{7}{11}\selectfont0.6 }\\\noalign{\vskip1pt}\hhline{~---------}\noalign{\vskip1pt}                                 &  TS &  \cellcolor{red!10}8.0\scalebox{0.6}{\ensuremath{\pm}}{\fontsize{7}{11}\selectfont0.8 } &  94.8\scalebox{0.6}{\ensuremath{\pm}}{\fontsize{7}{11}\selectfont0.8 } &  \cellcolor{red!10}6.8\scalebox{0.6}{\ensuremath{\pm}}{\fontsize{7}{11}\selectfont0.8 } &  20.3\scalebox{0.6}{\ensuremath{\pm}}{\fontsize{7}{11}\selectfont1.1 } &  \cellcolor{red!10}9.5\scalebox{0.6}{\ensuremath{\pm}}{\fontsize{7}{11}\selectfont1.0 } &  61.7\scalebox{0.6}{\ensuremath{\pm}}{\fontsize{7}{11}\selectfont1.6 } &  \cellcolor{red!10}15.8\scalebox{0.6}{\ensuremath{\pm}}{\fontsize{7}{11}\selectfont0.6 } &  37.4\scalebox{0.6}{\ensuremath{\pm}}{\fontsize{7}{11}\selectfont1.7 }\\\noalign{\vskip1pt}\hhline{~---------}\noalign{\vskip1pt}                                 &  Dirichlet &  \cellcolor{red!10}8.6\scalebox{0.6}{\ensuremath{\pm}}{\fontsize{7}{11}\selectfont0.6 } &  93.1\scalebox{0.6}{\ensuremath{\pm}}{\fontsize{7}{11}\selectfont1.6 } &  \cellcolor{red!10}7.7\scalebox{0.6}{\ensuremath{\pm}}{\fontsize{7}{11}\selectfont0.2 } &  20.9\scalebox{0.6}{\ensuremath{\pm}}{\fontsize{7}{11}\selectfont1.1 } &  \cellcolor{red!10}8.7\scalebox{0.6}{\ensuremath{\pm}}{\fontsize{7}{11}\selectfont0.9 } &  58.0\scalebox{0.6}{\ensuremath{\pm}}{\fontsize{7}{11}\selectfont1.4 } &  \cellcolor{red!10}16.3\scalebox{0.6}{\ensuremath{\pm}}{\fontsize{7}{11}\selectfont0.4 } &  33.1\scalebox{0.6}{\ensuremath{\pm}}{\fontsize{7}{11}\selectfont1.9 }\\\noalign{\vskip1pt}\hhline{~---------}\noalign{\vskip1pt}                                 &  SB &  \cellcolor{red!10}7.4\scalebox{0.6}{\ensuremath{\pm}}{\fontsize{7}{11}\selectfont0.8 } &  93.1\scalebox{0.6}{\ensuremath{\pm}}{\fontsize{7}{11}\selectfont2.7 } &  \cellcolor{red!10}5.9\scalebox{0.6}{\ensuremath{\pm}}{\fontsize{7}{11}\selectfont0.9 } &  17.9\scalebox{0.6}{\ensuremath{\pm}}{\fontsize{7}{11}\selectfont1.5 } &  \cellcolor{red!10}8.9\scalebox{0.6}{\ensuremath{\pm}}{\fontsize{7}{11}\selectfont1.1 } &  57.9\scalebox{0.6}{\ensuremath{\pm}}{\fontsize{7}{11}\selectfont3.9 } &  \cellcolor{red!10}15.0\scalebox{0.6}{\ensuremath{\pm}}{\fontsize{7}{11}\selectfont0.4 } &  35.5\scalebox{0.6}{\ensuremath{\pm}}{\fontsize{7}{11}\selectfont1.2 }\\\noalign{\vskip1pt}\hhline{~---------}\noalign{\vskip1pt}                                 &  Top-HB &  \cellcolor{red!10}7.7\scalebox{0.6}{\ensuremath{\pm}}{\fontsize{7}{11}\selectfont0.8 } &  94.1\scalebox{0.6}{\ensuremath{\pm}}{\fontsize{7}{11}\selectfont1.5 } &  \cellcolor{red!10}4.4\scalebox{0.6}{\ensuremath{\pm}}{\fontsize{7}{11}\selectfont0.5 } &  12.3\scalebox{0.6}{\ensuremath{\pm}}{\fontsize{7}{11}\selectfont0.4 } &  \cellcolor{red!10}8.8\scalebox{0.6}{\ensuremath{\pm}}{\fontsize{7}{11}\selectfont1.0 } &  58.8\scalebox{0.6}{\ensuremath{\pm}}{\fontsize{7}{11}\selectfont2.7 } &  \cellcolor{red!10}16.5\scalebox{0.6}{\ensuremath{\pm}}{\fontsize{7}{11}\selectfont0.5 } &  38.9\scalebox{0.6}{\ensuremath{\pm}}{\fontsize{7}{11}\selectfont1.6 }\\\noalign{\vskip1pt}\hhline{~---------}\noalign{\vskip1pt}                                 &  \textbf{Ours} &  \cellcolor{red!10}\textbf{4.5}\scalebox{0.6}{\ensuremath{\bm{\pm}}}{\fontsize{7}{11}\selectfont\textbf{1.4} } &  \textbf{95.6}\scalebox{0.6}{\ensuremath{\bm{\pm}}}{\fontsize{7}{11}\selectfont\textbf{1.3} } &  \cellcolor{red!10}\textbf{2.2}\scalebox{0.6}{\ensuremath{\bm{\pm}}}{\fontsize{7}{11}\selectfont\textbf{0.6} } &  \textbf{77.9}\scalebox{0.6}{\ensuremath{\bm{\pm}}}{\fontsize{7}{11}\selectfont\textbf{0.2} } &  \cellcolor{red!10}\textbf{1.8}\scalebox{0.6}{\ensuremath{\bm{\pm}}}{\fontsize{7}{11}\selectfont\textbf{1.2} } &  \textbf{81.3}\scalebox{0.6}{\ensuremath{\bm{\pm}}}{\fontsize{7}{11}\selectfont\textbf{0.5} } &  \cellcolor{red!10}\textbf{2.8}\scalebox{0.6}{\ensuremath{\bm{\pm}}}{\fontsize{7}{11}\selectfont\textbf{2.1} } &  \textbf{61.2}\scalebox{0.6}{\ensuremath{\bm{\pm}}}{\fontsize{7}{11}\selectfont\textbf{1.4} }\\\noalign{\vskip4pt} \hline \noalign{\vskip4pt}\noalign{\vskip1pt}\multirow{6}{*}{FMFP}                     & Softmax &  \cellcolor{red!10}4.8\scalebox{0.6}{\ensuremath{\pm}}{\fontsize{7}{11}\selectfont0.8 } &  84.2\scalebox{0.6}{\ensuremath{\pm}}{\fontsize{7}{11}\selectfont4.1 } &  \cellcolor{red!10}4.9\scalebox{0.6}{\ensuremath{\pm}}{\fontsize{7}{11}\selectfont0.4 } &  15.6\scalebox{0.6}{\ensuremath{\pm}}{\fontsize{7}{11}\selectfont1.7 } &  \cellcolor{red!10}5.4\scalebox{0.6}{\ensuremath{\pm}}{\fontsize{7}{11}\selectfont0.7 } &  45.4\scalebox{0.6}{\ensuremath{\pm}}{\fontsize{7}{11}\selectfont1.9 } &  \cellcolor{red!10}10.5\scalebox{0.6}{\ensuremath{\pm}}{\fontsize{7}{11}\selectfont0.3 } &  32.4\scalebox{0.6}{\ensuremath{\pm}}{\fontsize{7}{11}\selectfont1.4 }\\\noalign{\vskip1pt}\hhline{~---------}\noalign{\vskip1pt}                                 &  TS &  \cellcolor{red!10}8.0\scalebox{0.6}{\ensuremath{\pm}}{\fontsize{7}{11}\selectfont0.6 } &  95.3\scalebox{0.6}{\ensuremath{\pm}}{\fontsize{7}{11}\selectfont1.6 } &  \cellcolor{red!10}6.5\scalebox{0.6}{\ensuremath{\pm}}{\fontsize{7}{11}\selectfont0.3 } &  21.0\scalebox{0.6}{\ensuremath{\pm}}{\fontsize{7}{11}\selectfont1.5 } &  \cellcolor{red!10}9.5\scalebox{0.6}{\ensuremath{\pm}}{\fontsize{7}{11}\selectfont0.5 } &  57.7\scalebox{0.6}{\ensuremath{\pm}}{\fontsize{7}{11}\selectfont2.2 } &  \cellcolor{red!10}16.2\scalebox{0.6}{\ensuremath{\pm}}{\fontsize{7}{11}\selectfont1.1 } &  37.7\scalebox{0.6}{\ensuremath{\pm}}{\fontsize{7}{11}\selectfont1.8 }\\\noalign{\vskip1pt}\hhline{~---------}\noalign{\vskip1pt}                                 &  Dirichlet &  \cellcolor{red!10}8.2\scalebox{0.6}{\ensuremath{\pm}}{\fontsize{7}{11}\selectfont1.3 } &  94.0\scalebox{0.6}{\ensuremath{\pm}}{\fontsize{7}{11}\selectfont2.2 } &  \cellcolor{red!10}6.9\scalebox{0.6}{\ensuremath{\pm}}{\fontsize{7}{11}\selectfont0.4 } &  21.7\scalebox{0.6}{\ensuremath{\pm}}{\fontsize{7}{11}\selectfont1.2 } &  \cellcolor{red!10}8.9\scalebox{0.6}{\ensuremath{\pm}}{\fontsize{7}{11}\selectfont1.0 } &  56.6\scalebox{0.6}{\ensuremath{\pm}}{\fontsize{7}{11}\selectfont2.4 } &  \cellcolor{red!10}17.4\scalebox{0.6}{\ensuremath{\pm}}{\fontsize{7}{11}\selectfont0.8 } &  33.0\scalebox{0.6}{\ensuremath{\pm}}{\fontsize{7}{11}\selectfont1.8 }\\\noalign{\vskip1pt}\hhline{~---------}\noalign{\vskip1pt}                                 &  SB &  \cellcolor{red!10}7.2\scalebox{0.6}{\ensuremath{\pm}}{\fontsize{7}{11}\selectfont1.1 } &  93.1\scalebox{0.6}{\ensuremath{\pm}}{\fontsize{7}{11}\selectfont2.3 } &  \cellcolor{red!10}6.1\scalebox{0.6}{\ensuremath{\pm}}{\fontsize{7}{11}\selectfont0.5 } &  19.5\scalebox{0.6}{\ensuremath{\pm}}{\fontsize{7}{11}\selectfont1.0 } &  \cellcolor{red!10}8.6\scalebox{0.6}{\ensuremath{\pm}}{\fontsize{7}{11}\selectfont0.4 } &  55.8\scalebox{0.6}{\ensuremath{\pm}}{\fontsize{7}{11}\selectfont1.3 } &  \cellcolor{red!10}15.5\scalebox{0.6}{\ensuremath{\pm}}{\fontsize{7}{11}\selectfont0.6 } &  36.1\scalebox{0.6}{\ensuremath{\pm}}{\fontsize{7}{11}\selectfont0.5 }\\\noalign{\vskip1pt}\hhline{~---------}\noalign{\vskip1pt}                                 &  Top-HB &  \cellcolor{red!10}7.1\scalebox{0.6}{\ensuremath{\pm}}{\fontsize{7}{11}\selectfont0.6 } &  93.3\scalebox{0.6}{\ensuremath{\pm}}{\fontsize{7}{11}\selectfont4.9 } &  \cellcolor{red!10}5.2\scalebox{0.6}{\ensuremath{\pm}}{\fontsize{7}{11}\selectfont0.5 } &  14.2\scalebox{0.6}{\ensuremath{\pm}}{\fontsize{7}{11}\selectfont2.4 } &  \cellcolor{red!10}9.0\scalebox{0.6}{\ensuremath{\pm}}{\fontsize{7}{11}\selectfont0.7 } &  57.9\scalebox{0.6}{\ensuremath{\pm}}{\fontsize{7}{11}\selectfont2.4 } &  \cellcolor{red!10}16.2\scalebox{0.6}{\ensuremath{\pm}}{\fontsize{7}{11}\selectfont0.4 } &  37.4\scalebox{0.6}{\ensuremath{\pm}}{\fontsize{7}{11}\selectfont1.1 }\\\noalign{\vskip1pt}\hhline{~---------}\noalign{\vskip1pt}                                 &  \textbf{Ours} &  \cellcolor{red!10}\textbf{4.6}\scalebox{0.6}{\ensuremath{\bm{\pm}}}{\fontsize{7}{11}\selectfont\textbf{0.8} } &  \textbf{95.7}\scalebox{0.6}{\ensuremath{\bm{\pm}}}{\fontsize{7}{11}\selectfont\textbf{0.2} } &  \cellcolor{red!10}\textbf{3.0}\scalebox{0.6}{\ensuremath{\bm{\pm}}}{\fontsize{7}{11}\selectfont\textbf{0.4} } &  \textbf{77.4}\scalebox{0.6}{\ensuremath{\bm{\pm}}}{\fontsize{7}{11}\selectfont\textbf{0.2} } &  \cellcolor{red!10}\textbf{2.5}\scalebox{0.6}{\ensuremath{\bm{\pm}}}{\fontsize{7}{11}\selectfont\textbf{0.9} } &  \textbf{80.8}\scalebox{0.6}{\ensuremath{\bm{\pm}}}{\fontsize{7}{11}\selectfont\textbf{0.6} } &  \cellcolor{red!10}\textbf{1.8}\scalebox{0.6}{\ensuremath{\bm{\pm}}}{\fontsize{7}{11}\selectfont\textbf{2.0} } &  \textbf{60.8}\scalebox{0.6}{\ensuremath{\bm{\pm}}}{\fontsize{7}{11}\selectfont\textbf{1.4} }\\\noalign{\vskip4pt} \hline \noalign{\vskip4pt}\noalign{\vskip1pt}\multirow{6}{*}{Squentropy}                     & Softmax &  \cellcolor{red!10}\textbf{3.7}\scalebox{0.6}{\ensuremath{\bm{\pm}}}{\fontsize{7}{11}\selectfont\textbf{1.0} } &  88.2\scalebox{0.6}{\ensuremath{\pm}}{\fontsize{7}{11}\selectfont3.9 } &  \cellcolor{red!10}5.2\scalebox{0.6}{\ensuremath{\pm}}{\fontsize{7}{11}\selectfont0.5 } &  21.2\scalebox{0.6}{\ensuremath{\pm}}{\fontsize{7}{11}\selectfont1.8 } &  \cellcolor{red!10}4.6\scalebox{0.6}{\ensuremath{\pm}}{\fontsize{7}{11}\selectfont0.4 } &  52.0\scalebox{0.6}{\ensuremath{\pm}}{\fontsize{7}{11}\selectfont1.2 } &  \cellcolor{red!10}7.8\scalebox{0.6}{\ensuremath{\pm}}{\fontsize{7}{11}\selectfont0.3 } &  36.2\scalebox{0.6}{\ensuremath{\pm}}{\fontsize{7}{11}\selectfont0.8 }\\\noalign{\vskip1pt}\hhline{~---------}\noalign{\vskip1pt}                                 &  TS &  \cellcolor{red!10}6.2\scalebox{0.6}{\ensuremath{\pm}}{\fontsize{7}{11}\selectfont1.1 } &  95.6\scalebox{0.6}{\ensuremath{\pm}}{\fontsize{7}{11}\selectfont0.9 } &  \cellcolor{red!10}6.9\scalebox{0.6}{\ensuremath{\pm}}{\fontsize{7}{11}\selectfont0.6 } &  28.2\scalebox{0.6}{\ensuremath{\pm}}{\fontsize{7}{11}\selectfont2.5 } &  \cellcolor{red!10}8.3\scalebox{0.6}{\ensuremath{\pm}}{\fontsize{7}{11}\selectfont0.6 } &  66.6\scalebox{0.6}{\ensuremath{\pm}}{\fontsize{7}{11}\selectfont1.4 } &  \cellcolor{red!10}13.3\scalebox{0.6}{\ensuremath{\pm}}{\fontsize{7}{11}\selectfont0.1 } &  44.9\scalebox{0.6}{\ensuremath{\pm}}{\fontsize{7}{11}\selectfont1.0 }\\\noalign{\vskip1pt}\hhline{~---------}\noalign{\vskip1pt}                                 &  Dirichlet &  \cellcolor{red!10}6.5\scalebox{0.6}{\ensuremath{\pm}}{\fontsize{7}{11}\selectfont1.2 } &  95.9\scalebox{0.6}{\ensuremath{\pm}}{\fontsize{7}{11}\selectfont0.8 } &  \cellcolor{red!10}7.3\scalebox{0.6}{\ensuremath{\pm}}{\fontsize{7}{11}\selectfont0.3 } &  29.4\scalebox{0.6}{\ensuremath{\pm}}{\fontsize{7}{11}\selectfont1.1 } &  \cellcolor{red!10}7.8\scalebox{0.6}{\ensuremath{\pm}}{\fontsize{7}{11}\selectfont0.6 } &  64.0\scalebox{0.6}{\ensuremath{\pm}}{\fontsize{7}{11}\selectfont1.3 } &  \cellcolor{red!10}14.1\scalebox{0.6}{\ensuremath{\pm}}{\fontsize{7}{11}\selectfont0.3 } &  42.5\scalebox{0.6}{\ensuremath{\pm}}{\fontsize{7}{11}\selectfont0.7 }\\\noalign{\vskip1pt}\hhline{~---------}\noalign{\vskip1pt}                                 &  SB &  \cellcolor{red!10}6.0\scalebox{0.6}{\ensuremath{\pm}}{\fontsize{7}{11}\selectfont0.8 } &  95.3\scalebox{0.6}{\ensuremath{\pm}}{\fontsize{7}{11}\selectfont1.2 } &  \cellcolor{red!10}6.2\scalebox{0.6}{\ensuremath{\pm}}{\fontsize{7}{11}\selectfont0.4 } &  23.8\scalebox{0.6}{\ensuremath{\pm}}{\fontsize{7}{11}\selectfont1.9 } &  \cellcolor{red!10}7.8\scalebox{0.6}{\ensuremath{\pm}}{\fontsize{7}{11}\selectfont0.7 } &  63.0\scalebox{0.6}{\ensuremath{\pm}}{\fontsize{7}{11}\selectfont2.9 } &  \cellcolor{red!10}13.0\scalebox{0.6}{\ensuremath{\pm}}{\fontsize{7}{11}\selectfont0.5 } &  45.2\scalebox{0.6}{\ensuremath{\pm}}{\fontsize{7}{11}\selectfont2.0 }\\\noalign{\vskip1pt}\hhline{~---------}\noalign{\vskip1pt}                                 &  Top-HB &  \cellcolor{red!10}5.3\scalebox{0.6}{\ensuremath{\pm}}{\fontsize{7}{11}\selectfont0.4 } &  96.4\scalebox{0.6}{\ensuremath{\pm}}{\fontsize{7}{11}\selectfont0.9 } &  \cellcolor{red!10}4.3\scalebox{0.6}{\ensuremath{\pm}}{\fontsize{7}{11}\selectfont0.5 } &  15.8\scalebox{0.6}{\ensuremath{\pm}}{\fontsize{7}{11}\selectfont1.4 } &  \cellcolor{red!10}8.2\scalebox{0.6}{\ensuremath{\pm}}{\fontsize{7}{11}\selectfont0.8 } &  66.5\scalebox{0.6}{\ensuremath{\pm}}{\fontsize{7}{11}\selectfont2.2 } &  \cellcolor{red!10}13.7\scalebox{0.6}{\ensuremath{\pm}}{\fontsize{7}{11}\selectfont0.1 } &  45.9\scalebox{0.6}{\ensuremath{\pm}}{\fontsize{7}{11}\selectfont1.4 }\\\noalign{\vskip1pt}\hhline{~---------}\noalign{\vskip1pt}                                 &  \textbf{Ours} &  \cellcolor{red!10}4.1\scalebox{0.6}{\ensuremath{\pm}}{\fontsize{7}{11}\selectfont0.8 } &  \textbf{97.2}\scalebox{0.6}{\ensuremath{\bm{\pm}}}{\fontsize{7}{11}\selectfont\textbf{0.5} } &  \cellcolor{red!10}\textbf{2.3}\scalebox{0.6}{\ensuremath{\bm{\pm}}}{\fontsize{7}{11}\selectfont\textbf{0.5} } &  \textbf{79.0}\scalebox{0.6}{\ensuremath{\bm{\pm}}}{\fontsize{7}{11}\selectfont\textbf{0.3} } &  \cellcolor{red!10}\textbf{3.3}\scalebox{0.6}{\ensuremath{\bm{\pm}}}{\fontsize{7}{11}\selectfont\textbf{0.8} } &  \textbf{82.9}\scalebox{0.6}{\ensuremath{\bm{\pm}}}{\fontsize{7}{11}\selectfont\textbf{0.4} } &  \cellcolor{red!10}\textbf{0.6}\scalebox{0.6}{\ensuremath{\bm{\pm}}}{\fontsize{7}{11}\selectfont\textbf{0.2} } &  \textbf{66.5}\scalebox{0.6}{\ensuremath{\bm{\pm}}}{\fontsize{7}{11}\selectfont\textbf{0.7} }\\\noalign{\vskip4pt} \bottomrule\noalign{\vskip1pt}
    \end{tabular}
    }
    \vspace{3pt}
    \caption{In every round the error was enforced to be below 5\%; `TS' stands for Temperature Scaling, `SB' stands for Scaling Binning, `Top-HB' stands for Top-Label Histogram Binning. The column Err stands for auto-labeling error and Cov stands for the coverage. Each cell value is mean $\pm$ std. deviation observed on 5 repeated runs with different random seeds. }
    \label{table:final_table}
    \end{table*}
    


\vspace{-0pt}
\subsubsection{Post-hoc methods}
We use the following methods for learning (or updating) the confidence function $\hat{\fg}$ after learning $\hat{\fh}$.

\begin{enumerate}[topsep=0pt,itemsep=0ex,partopsep=1ex,parsep=1ex,leftmargin=24pt]
    \item 
\textit{Temperature scaling} ~\citep{guo2017calibration} is a variant of Platt scaling ~\citep{platt1999SVM-Prob}. It rescales the logits by a learnable scalar parameter. 

\item \textit{Top-Label Histogram-Binning} ~\citep{gupta2021top} builds on the histogram-binning method \citep{zadrozny2002transforming} and focuses on calibrating the scores of the predicted label assigned to unlabeled points.
 
\item 
\textit{Scaling-Binning}
~\citep{kumar2019verified} applies temperature scaling and then bins the confidence function values.

\item \textit{Dirichlet Calibration} ~\citep{kull2019Dirichlet}
models the distribution of predicted probability vectors separately on instances of each class and assumes Dirichlet class conditional distributions.





\end{enumerate}

\textbf{Remark:} Each train-time method is piped with a post-hoc method, yielding total $4 \times 5 = 20$ methods.

\subsection{Datasets and models} \label{sec:dataset_and_model}
    We evaluate the performance of auto-labeling on four datasets. Each is paired with a model for auto-labeling:
    \begin{enumerate}[topsep=0pt,itemsep=0.5ex,partopsep=1ex,parsep=1ex,leftmargin=18pt]
        \item \emph{MNIST} \cite{lecun1998mnist} is a hand-written digits dataset. We use the LeNet \cite{lecun1998LeNet} for auto-labeling.
        
        \item \emph{CIFAR-10} \cite{krizhevsky2009CIFAR-10} is an image dataset with 10 classes. We use a CNN with approximately 5.8M parameters \cite{mednet} for auto-labeling.
        
        \item \emph{Tiny-ImageNet} \cite{le2015tiny} is an image dataset comprising 100K images across 200 classes. 
        We use CLIP \cite{radford2021learning} to derive embeddings for the images in the dataset 
        and use an MLP model.
        \item \emph{20 Newsgroups} \cite{misc_twenty_newsgroups_113} is a natural language dataset comprising around 18K news posts across 20 topics. We use the FlagEmbedding \cite{bge_embedding} to obtain text embeddings and use an MLP model.

    \end{enumerate}

\subsection{Hyperparameter Search and Evaluation}
\label{subsec:hyp-search}
 The complexity of TBAL workflow and lack of labeled data make hyperparameter search and evaluation challenging. Similar challenges have been observed in active learning ~\citep{lowell2019ObstaclesAL}. We discuss our practical approach and defer the details to Appendix \ref{subsec:exp-protocol}.




\textbf{Hyperparameter Search.}  We run only the first round of TBAL with each method using a hyperparameter combination 5 times and measure the mean auto-labeling error and mean coverage on $D_{\mathrm{hyp}}$, which represents a small part of the held-out human-labeled data. We pick the combination that yields the lowest average auto-labeling error while maximizing the coverage.
We first find the best hyperparameters for each train-time method, fix those, and then search the hyperparameters for the post-hoc methods. Note that the best hyperparameter for a post-hoc method depends on the training-time method that it pipes to.
The hyperparameter search spaces are in the Appendix \ref{sec:experiments_details}; and the selected values used for each setting are in the supplementary material.
%


\textbf{Performance Evaluation}. After fixing the hyper-parameters, we run TBAL with each combination of train-time and post-hoc method on full $X_u$ of size $N$, with a fixed budget of $N_t$ labeled training samples and $N_v$ validation samples. The details of these values for each dataset are in Table \ref{tab:settings_details} in Appendix \ref{sec:experiments_details}. Here, we know the ground truth labels for the points in $X_u$, so we measure the auto-labeling error and coverage as defined in equations \eqref{eq:auto-err} and \eqref{eq:auto-cov} respectively and report them in Table \ref{table:final_table}. We discuss these results and their implications in the next section.


\vspace{-0pt}
\subsection{Results and Discussion}
\vspace{-0pt}
Our findings are, shown in Table \ref{table:final_table}, are:

\textbf{\textit{C1: \ourmethod\, improves TBAL performance.}} Our approach aims to optimize the confidence function to maximize coverage while minimizing errors. When applied to TBAL, we expect it to yield substantial coverage enhancement and error reduction compared to vanilla training and softmax scores. Indeed, the results in Table \ref{table:final_table} corresponding to the vanilla training match our expectations. We see \emph{across all data settings}, our method achieves \textit{significantly higher coverage} while keeping auto-labeling error below the tolerance level of 5\%. The improvements are even more pronounced when the datasets are more complex than MNIST. Also consistent with our expectation and observations in Figure \ref{fig:calib_scores}, the post-hoc calibration methods improve the coverage over using softmax scores but at the cost of slightly higher error. While they are reasonable choices to apply in the TBAL pipeline, they fall short of maximally improving TBAL performance due to the misalignment of goals. 

\textbf{\textit{C2: \ourmethod\, is compatible with and improves over other train-time methods.}}
Our method is compatible with various choices of train-time methods, and if a train-time method (Squentropy here) provides a \emph{better} model relative to another train-time method (e.g., Vanilla), then our method exploits this gain and pushes the performance even further. Across different train-time methods, we do not see significant differences in the performance, except for Squentropy. Using Squentropy with softmax improves the coverage by as high as 6-7\% while dropping the auto-labeling error in contrast to using softmax scores obtained with other train-time methods for the Tiny-ImageNet setting. 
This is an unexpected and interesting finding. Squentropy adds average square loss over the incorrect classes as a regularizer, and it has been shown to achieve better accuracy and calibration compared to training just with cross-entropy loss (Vanilla). 


\textbf{\textit{Train-time methods designed for ordinal ranking objective perform poorly in auto-labeling.}}
CRL and FMFP are state-of-the-art methods designed to produce scores aligned with the ordinal ranking criteria. Ideally, if the scores satisfy this criterion, TBAL's performance would improve. However, we do not see any significant difference from the Vanilla method. Similar to the other baselines, their evaluation is focused on models trained on large amounts of data. But, in TBAL, we have less data for training. The training error goes to zero after some rounds, and no information is left for the CRL loss to distinguish between correct and incorrect predictions (i.e., count SGD mistakes). On the other hand, FMFP is based on a hypothesis that training models using Sharpness Aware Minimizer (SAM) could lead to scores satisfying the ordinal ranking criteria. However, this phenomenon is still not well understood, especially in settings like ours with limited training data.



\vspace{-0pt}
\section{Related Works}


\textbf{Data-labeling.} We briefly discuss prominent methods for data labeling. 
Crowdsourcing \cite{raykar_learning_from_crowds, sorokin_annotation} uses a crowd of non-experts to complete a set of labeling tasks. Works in this domain focus on mitigating noise in the obtained information, modeling label errors, and designing effective labeling tasks \cite{Ryan2011, karger2011budget, mazumdarsaha, vinayak2014graph, vinayak_triangle, vinayak2017tensor, chen2023crowdsourced}. 
Weak supervision, in contrast, emphasizes labeling through multiple inexpensive but noisy sources, not necessarily human \citep{Ratner16, fu2020fast, shin2022universalizing, Vishwakarma2022Lifting}. Works such as \citet{Ratner16, fu2020fast} concentrate on binary or multi-class labeling, while \citet{shin2022universalizing,Vishwakarma2022Lifting} extend weak supervision to structured prediction tasks.

Auto-labeling occupies an intermediate position between weak supervision and crowdsourcing in terms of human dependency. It aims to minimize costs associated with obtaining labels from humans while generating high-quality labeled data using a specific machine learning model. \citet{qiu2020min-cost-al} use a TBAL-like algorithm and explore the cost of training for auto-labeling with large-scale model classes. Recent work \cite{vishwakarma2023promises} theoretically analyzes the sample complexity of validation data required to guarantee the quality of auto-labeled data.

\textbf{Overconfidence and calibration.} 
The issue of overconfidence ~\citep{szegedy2014Intriguing, nguyen2015deep, hein2018Confidence, bai2021OverConfLogisticReg}
is detrimental in several applications, including ours. Many solutions have emerged to mitigate the overconfidence and miscalibration problem. ~\citet{gawlikowski2021survey} provide a comprehensive survey on uncertainty quantification and calibration techniques for neural networks. \citet{guo2017calibration} evaluated a variety of solutions ranging from the choice of network architecture, model capacity, weight decay regularization ~\citep{weightdecay1991}, histogram-binning and isotonic regression ~\citep{zadrozny2001learning,zadrozny2002transforming} and temperature scaling ~\citep{platt1999SVM-Prob, mizil2005Scaling} which they found to be the most promising solution. The solutions fall into two broad categories: train-time and post-hoc. Train-time solutions modify the loss function, include additional regularization terms, or use different training procedures ~\citep{aviral2018MMCE, muller2019does, mukhoti2020focal, hui2023Squentropy}.
%
On the other hand, post-hoc methods such as top-label histogram-binning~\citep{gupta2021distribution}, scaling binning ~\citep{kumar2019verified}, Dirichlet calibration ~\citep{kull2019Dirichlet} calibrate the scores directly or learn a model that corrects miscalibrated confidence scores.

\textbf{Beyond calibration.}
While calibration aims to match the confidence scores with a probability of correctness, it is not the precise solution to the overconfidence problem in many applications, including our setting. The desirable criteria for scores for TBAL are closely related to the ordinal ranking criterion ~\citep{hendrycks2017OrdinalOOD}. To get such scores, ~\citet{corbi2019Confidnet} add an additional module in the net for failure prediction,
 \citet{zhu2022FMFP} switch to sharpness aware minimization \cite{foret2021sharpnessaware} to learn the model; CRL  \citep{moon2020CRL} regularizes the loss function.

%


\vspace{0pt}
\section{Conclusions}
We studied issues with confidence scoring functions used in threshold-based auto-labeling (TBAL). We showed that the commonly used confidence functions and calibration methods can often be a bottleneck, leading to poor performance. We proposed \ourmethod\,  to learn confidence functions that are aligned with the TBAL objective. We evaluated our method extensively against common baselines on several real-world datasets and found that it improves the performance of TBAL significantly in comparison to the several common choices of confidence function. Our method is compatible with several choices of methods used for training the classifier in TBAL and using it in conjunction with them improves TBAL performance further. A limitation of \ourmethod\, is that, similar to other post-hoc methods it also requires validation data to learn the confidence function. Reducing (or eliminating) this dependence on validation data could be an interesting future work.

\section{Acknowledgments}
This work was partly supported by funding from the American Family Data Science Institute. We thank Heguang Lin, Changho Shin, Dyah Adila, Tzu-Heng Huang, John Cooper, Aniket Rege, Daiwei Chen and  Albert Ge for their valuable inputs.  We thank the anonymous reviewers for their valuable comments and constructive feedback on our work.




\bibliography{references}

\bibliographystyle{abbrvnat}



\newpage 

\appendix

\section{Supplementary Material Organization}
The supplementary material is organized as follows. We provide deferred details of the method in section \ref{sec:method_details}. Then, in section \ref{sec:experiments_details}, we provide additional experimental results and details of the experiment protocol and hyperparameters used for the experiments. Our code with instructions to run, is uploaded along with the paper.

\subsection{Glossary} \label{appendix:glossary}
\label{sec:gloss}
The notation is summarized in Table~\ref{table:glossary} below. 
\begin{table*}[h]
\centering
\begin{tabular}{l l}
\toprule
Symbol & Definition \\
\midrule
$\one(E)$ & indicator function of event $E$. It is $1$ if $E$
happens and $0$ otherwise.\\
$\mathcal{X}$ & feature space. \\
$\mathcal{Y}$ & label space i.e. ${1,2,\ldots k}$.\\
$\cH$ & hypothesis space (model class for the classifiers). \\
$\cG$ & class of confidence functions.\\
$k$ & number of classes.\\

$\bx,y$ & $\bx$ is an element in $\cX$  and $y$ is its true label.\\ 

$\fh$ & a hypothesis (model) in $\cH$.\\
$\fg$ & confidence function $\fg:\cX \to \Delta^{k}$.\\

$X_{u}$ & given pool of unlabeled data points.\\
$X_{u}^{(i)}$ & unlabeled data left at the beginning of $i$th round.  \\
$\hat{\fh}^{(i)}$ & ERM solution and auto-labeling thresholds respectively in $i$th round.\\
$D_\mathrm{query}^{(i)}$ & labeled data queried from oracle (human) in the $i$th round.\\ \vspace{3pt}
$D_\mathrm{train}^{(i)}$ & training data to learn $\hat{\fh}^{(i)}$ in the $i$th round.\\ \vspace{3pt}
$D_\mathrm{val}^{(i)}$ & validation data in the $i$th round.\\ \vspace{3pt}
$D_\mathrm{cal}^{(i)}$ & calibration data in the $i$th round 
 to learn a post-hoc $\fg$.\\ \vspace{3pt}
$D_\mathrm{th}^{(i)}$ &  part of validation data in the $i$th round to estimate threshold $\vt$.\\ \vspace{3pt}
$D_\mathrm{auto}^{(i)}$ &  part of $X_u^{(i)}$ that got auto-labeled in the $i$th round.\\ \vspace{3pt}
$D_\mathrm{out}$ & Output labeled data, including auto-labeled and human labeled data.\\

$\vt$ & $k$ dimensional vector of thresholds.\\
$\vt[y]$ & $y$th entry of $\vt$ i.e. the threshold for class $y$. \\
$\fg(\bx)[y]$ & the confidence score for class $y$ output by confidence function $\fg$ on data point $\bx$. \\
$\hat{y}$ & predicted class for data point $\bx$.\\
$\ff^*$ & unknown groundtruth labeling function.\\ 
$N_u$ & number of unlabeled points, i.e. size of $X_u$.\\
$N_t$ & number of manually labeled points that can be used for training $\fh$.\\
$N_a$ & Total auto-labeled points in $D_{\mathrm{out}}$. \\
$\nu$ & fraction of $\Dval$ that can be used for training post-hoc calibrator. \\
$A$ & indices of points that are auto-labeled. \\
$X_u(A)$ & subset of points in $X_u$ with indices in $A$, i.e. the set of auto-labeled points. \\
$\tilde{y}_i$ & label assigned to the $i$th point by the algorithm. It could be either $y_i$ or $\hat{y}_i$. \\
$y_i$ & groundtruth label for the $i$th point. \\
$\hat{y}_i$ & predicted label for the $i$th point by classifier.\\
$\epsilon_a$ & auto-labeling error tolerance.\\

$\err (\fg, \vt \given \fh)$ & population level auto-labeling error, see eq. \eqref{eq:true_err}.\\
$\cov (\fg, \vt \given \fh)$ & population level auto-labeling coverage, see eq. \eqref{eq:true_cov}.\\

$\haterr (\fg, \vt \given \fh, D)$ & estimated auto-labeling error, see eq. \eqref{eq:est_err}.\\
$\hatcov (\fg, \vt \given \fh, D)$ & estimated auto-labeling coverage, see eq. \eqref{eq:est_cov}.\\

$\surrerr (\fg, \vt \given \fh, D)$ & surrogate estimated auto-labeling error, see eq. \eqref{eq:surr_err}.\\
$\surrcov (\fg, \vt \given \fh, D)$ & surrogate estimated auto-labeling coverage, see eq. \eqref{eq:surr_cov}.\\
 
\toprule
\end{tabular}
\caption{
	Glossary of variables and symbols used in this paper.
}
\label{table:glossary}
\end{table*}

\newpage

\section{ Appendix to Section \ref{section:OurMethod}} \label{sec:method_details}
\begin{wrapfigure}{r}{0.6\linewidth}
  
    \mbox{
    \subfigure[ Auto-labeling error and surrogate error at various $\alpha$.\label{fig:surr_err}]{\includegraphics[width=\linewidth]{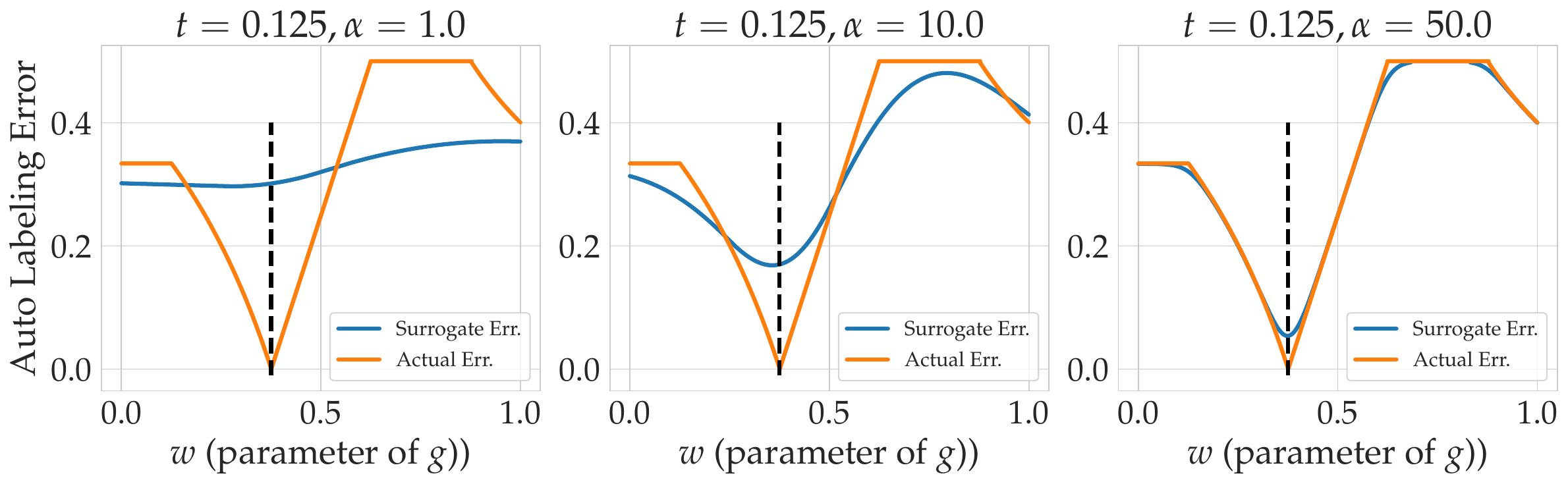}}
  }
    \mbox{
    \subfigure[ Auto-labeling coverage and surrogate coverage at various $\alpha$.\label{fig:surr_cov}]{\includegraphics[width=\linewidth]{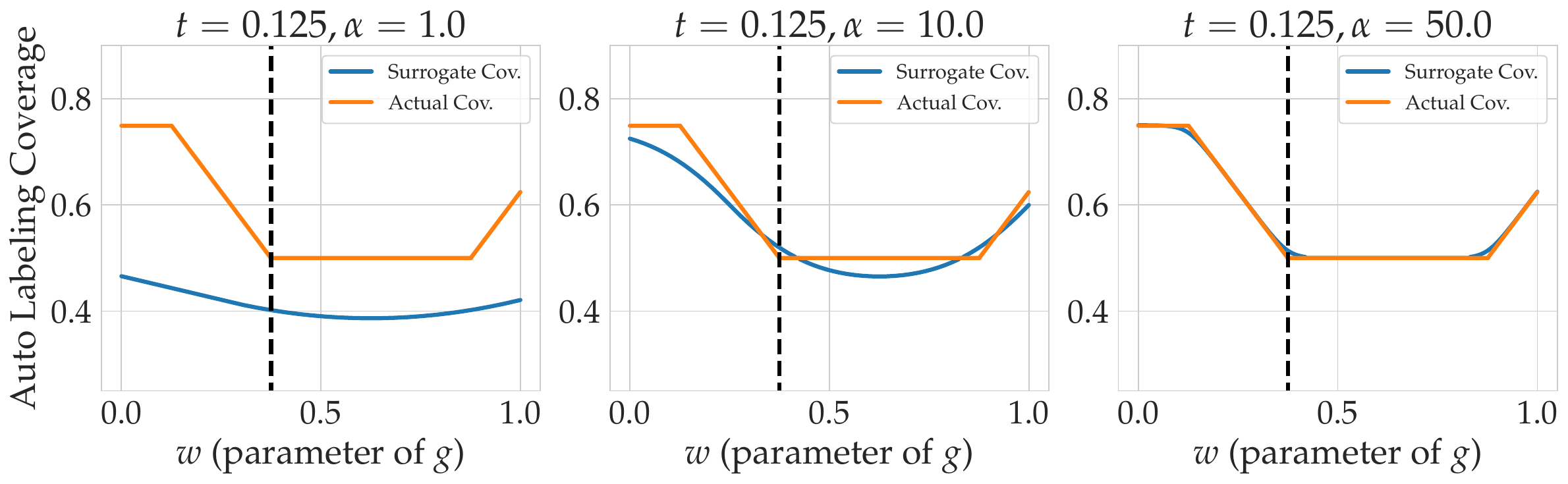}}
  }
  
  \caption{Illustration of the tightness of surrogate error and coverage functions based on the choice of $\alpha$.}
  \label{fig:example-key}
\end{wrapfigure}

\textbf{Tightness of surrogates.} The surrogate auto-labeling error and coverage introduced to relax the optimization problem \eqref{eq:emp_opt} is indeed a good approximation of the actual auto-labeling error and coverage. To see this, we use a toy data setting of $x \sim \mathrm{Uniform}(0,1)$ with $1-$dimensional threshold classifier $\fh_{\theta}(x) = \one(x\ge \theta)$. For any $x$, let true labels $y=\fh_{0.5}(x)$ and consider the confidence function $\fg_w(x)=|w-x|$. Let $\hat{y} = \fh_{0.25}(x)$ and consider the points on the side where $\hat{y}=1$. We plot actual and surrogate errors in Figure \ref{fig:surr_err} and the surrogate and actual coverage in Figure \ref{fig:surr_err}.


 for three choices of $\alpha$. As expected, the gap between the surrogates and the actual functions diminishes as we increase the $\alpha$.

\textbf{Active Querying Strategy.} We employ the margin-random query approach to select the next batch of training data. This method involves sorting points based on their margin (uncertainty) scores and selecting the top $Cn_b$ points, from which $n_b$ points are randomly chosen. This strategy provides a straightforward and computationally efficient way to balance the exploration-exploitation trade-off. It's important to acknowledge the existence of alternative active-querying strategies; however, we adopt the margin-random approach as our standard to maintain a focus on evaluating various choices of confidence functions for auto-labeling. Note that while we use the new confidence scores computed using post-hoc methods for auto-labeling, we do not use these scores in active querying. Instead, we use the softmax scores from the model for this. We do this to avoid conflating the study with the study of active querying strategies. We use $C=2$ for all experiments.


\newpage
\newpage
\clearpage
\section{ Additional Experiments and Details} \label{sec:experiments_details}
\subsection{Details of the experiment in section 2.2}
\label{subsec:motivation_exp_details}

We run TBAL for a single round on the CIFAR-10 dataset with a SimpleCNN classification model with around 5.8M parameters ~\citep{mednet}. We randomly sampled 4,000 points for training the classifier and randomly sampled 1,000 points as validation data. We train the model to zero training error using minibatch SGD with learning rate 1e-3, weight decay 1e-3 \cite{weightdecay1988, weightdecay1991}, momentum 0.9, and batch size 32. The trained model has validation accuracy around 55\%, implying we could hope to get coverage around 55\%. We run the auto-labeling procedure with an error tolerance of 5\%.

\subsection{Experiments on $N_t$, $N_v$ and $\nu$}


\begin{figure*}[h]
  \centering
 \begin{subfigure}
    \centering
    \includegraphics[width=1.0\textwidth]{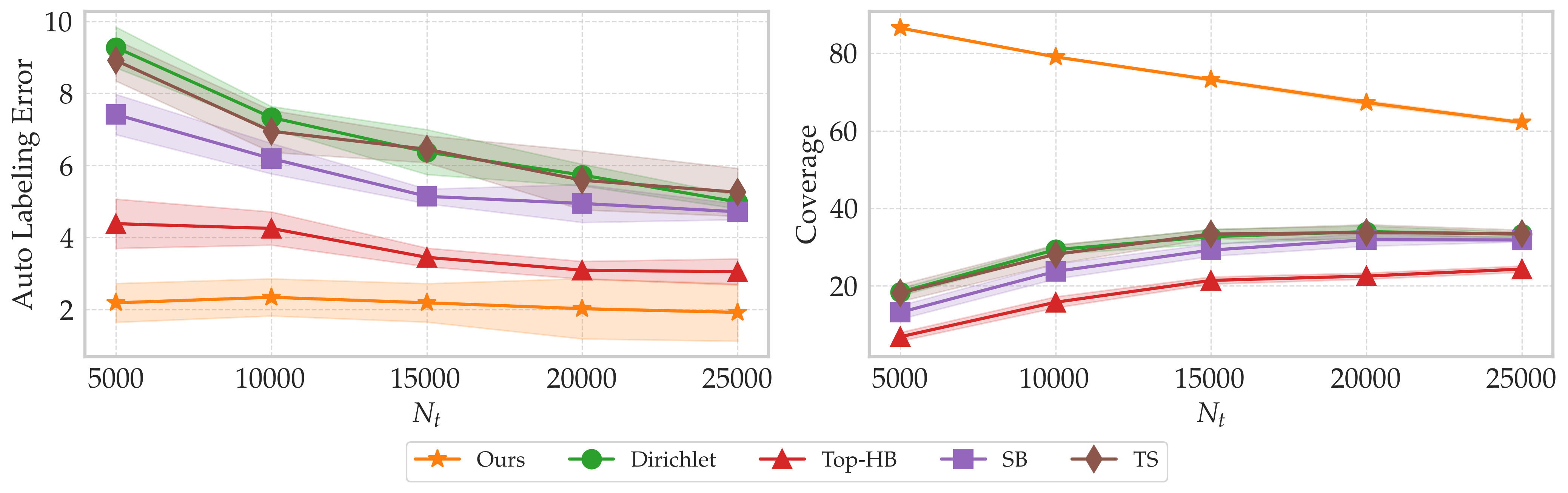}
    \caption{\normalsize Autolabeling error and coverage of different post-hoc methods on CIFAR-10 for various $N_t$}
    \label{fig:nt}
  \end{subfigure}  

   \begin{subfigure}
    \centering
    \includegraphics[width=1.0\textwidth]{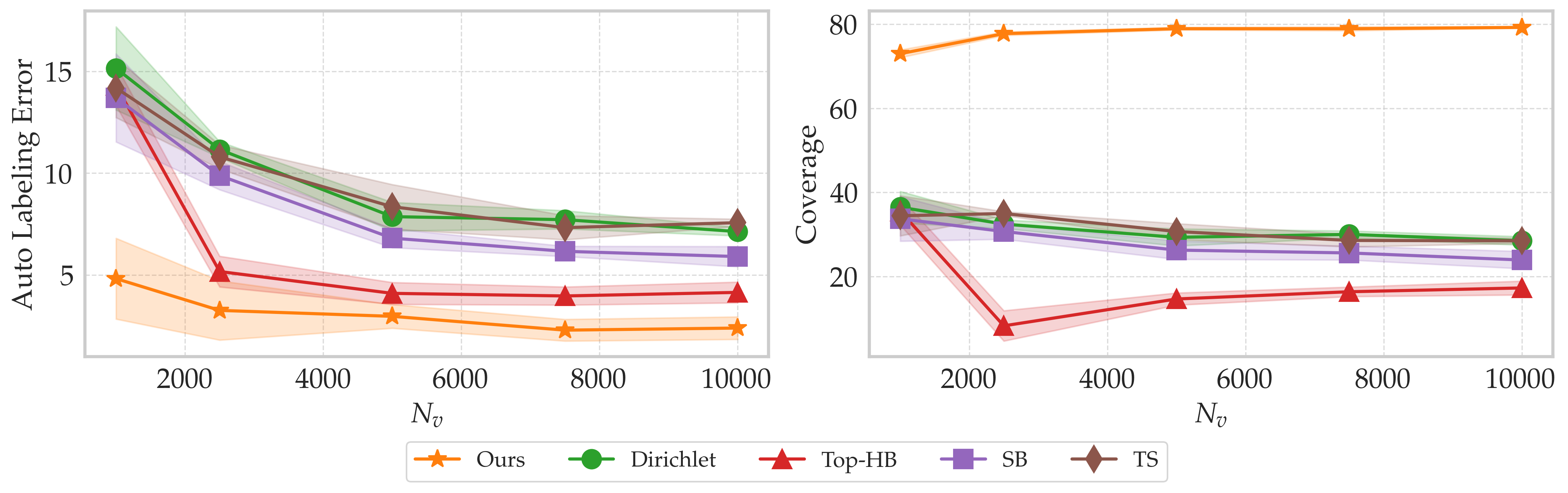}
    \caption{\normalsize Autolabeling error and coverage of different post-hoc methods on CIFAR-10 for various $N_v$}
    \label{fig:nv}
  \end{subfigure} 

   \begin{subfigure}
    \centering
    \includegraphics[width=1.0\textwidth]{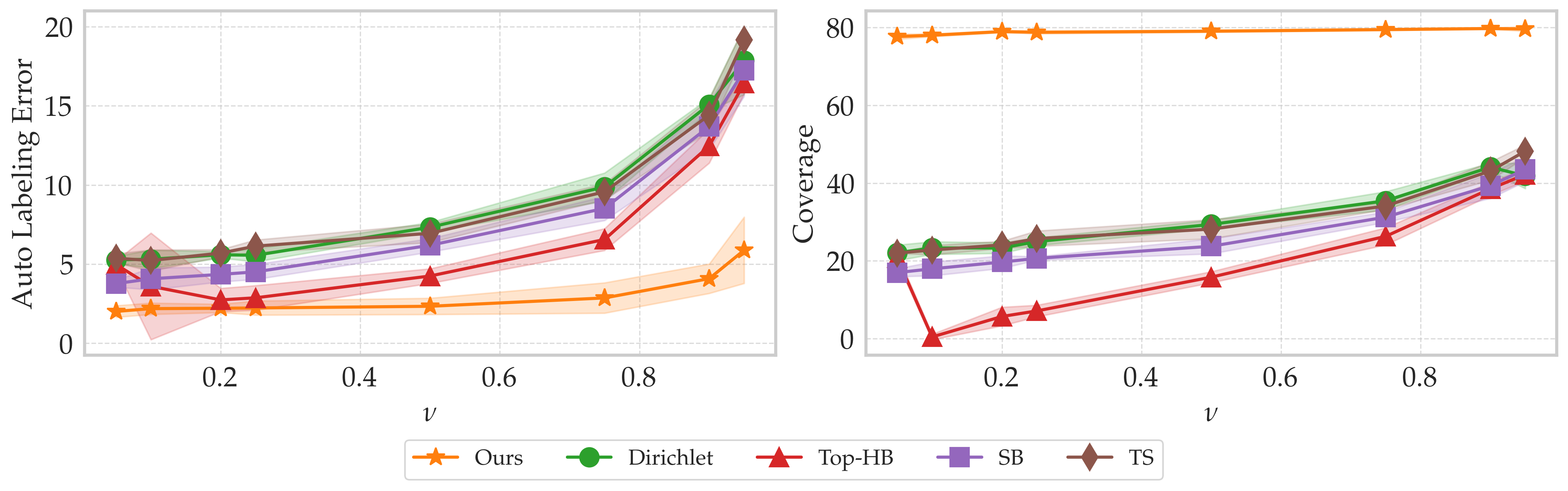}
    \caption{\normalsize Autolabeling error and coverage of different post-hoc methods on CIFAR-10 for various $\nu$}
    \label{fig:calib_val_frac}
  \end{subfigure} 
  \label{fig:three_graphs}
\end{figure*}
  
We need to understand the effect of training data query budget i.e. $N_t$, the total validation data $N_v$, and the data that can be used for calibrating the model i.e. the calibration data fraction $\nu$ on the auto-labeling objective. As varying these hyperparameters on each train-time method is expensive, we experimented with only Squentropy as it was the best-performing method across settings for various datasets. 

When we vary the budget for training data $N_t$, we observe from Figure \ref{fig:nt} that our method does not require a lot of data to train the base model,  i.e. achieving low auto-labeling error and high coverage with a low budget. While other methods benefit from having more training data for auto-labeling objectives, it comes at the expense of reducing the available data for validation.

From figure \ref{fig:nv}, we observe that, while the coverage of our method remains the same across different $N_v$, it reduces for other methods. The cause of this phenomenon can be attributed to the fact that we are borrowing the data from the training budget as it limits the performance of the base model, which in turn limits the auto-labeling objective.  

As we increase the percentage of data that can be used to calibrate the model, i.e., $\nu$, we note from figure \ref{fig:calib_val_frac} that other methods improve the coverage, which can be understood from the fact that when more data is available for calibrating the model, the model becomes better in terms of the auto-labeling objective. But it's interesting to note that even with a low calibration fraction, our method achieves superior coverage compared to other methods. It is also important to note that the auto-labeling error increases as we increase $\nu$. This is because when $\nu$ increases, the number of data points used to estimate the threshold decreases, leading to a less granular and precise threshold.

\begin{figure*}[h]
  \centering
  \begin{subfigure}
    \centering
    \includegraphics[width=1.0 \textwidth]{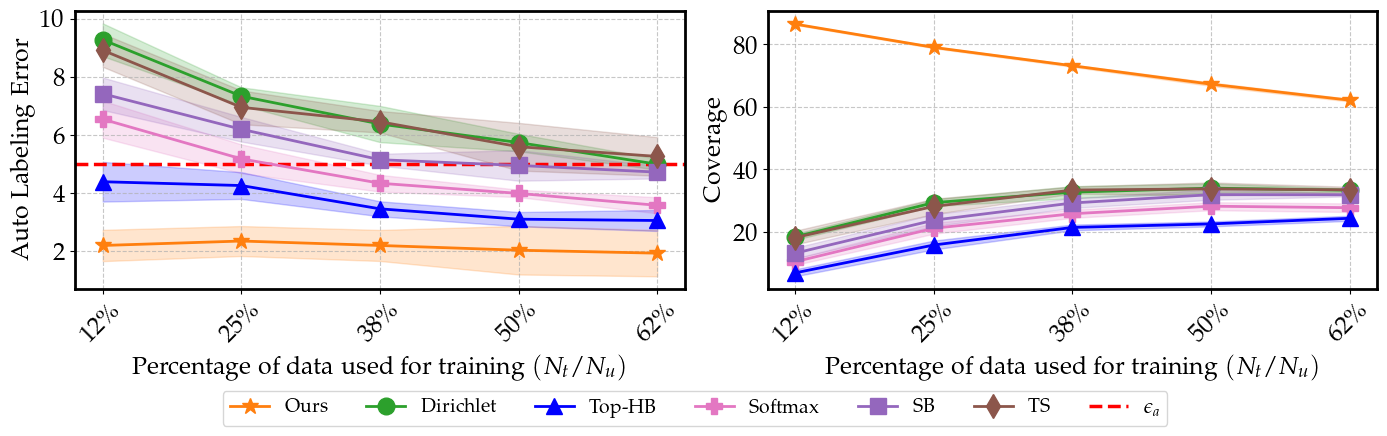}
    \caption{\normalsize Auto-labeling error and coverage for different post-hoc methods on CIFAR-10 while we vary $N_t$. $N_u=40,000$ is the size of the given unlabeled pool.}
    \label{fig:nt-cifar10}
  \end{subfigure}  
  
  \begin{subfigure}
    \centering
    \includegraphics[width=1.0\textwidth]{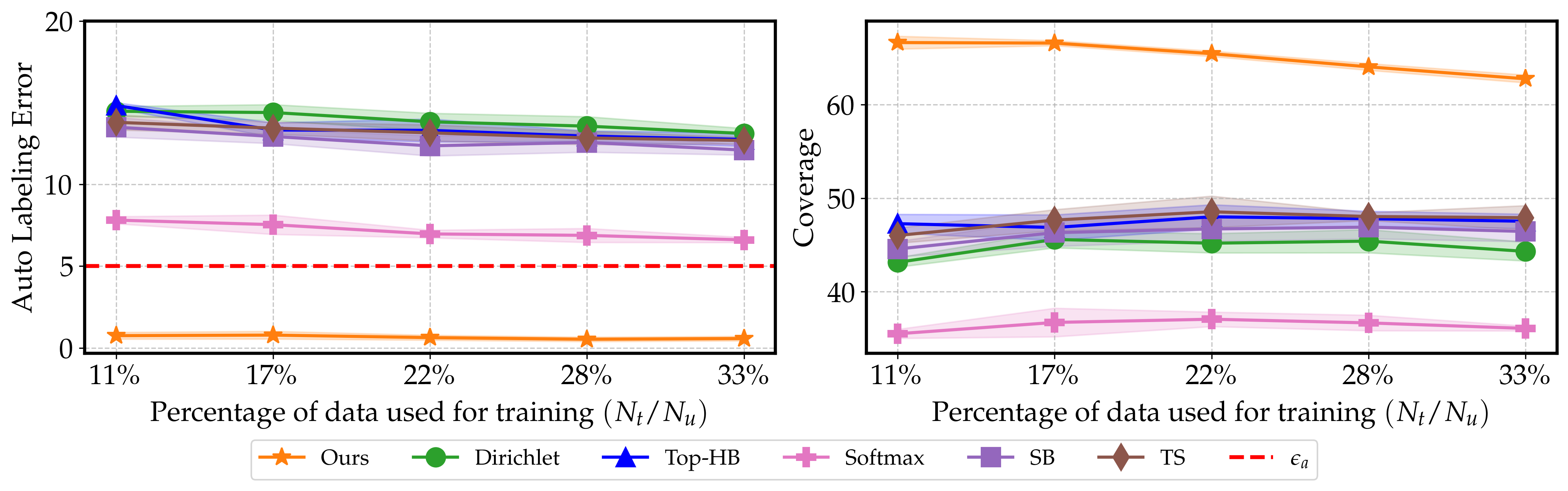}
    \caption{\normalsize Auto-labeling error and coverage for different post-hoc methods on Tiny-ImageNet while we vary $N_t$. $N_u=90,000$ is the size of the given unlabeled pool.}
    \label{fig:nt-tinyimagenet}
  \end{subfigure}  
  
  \begin{subfigure}
    \centering
    \includegraphics[width=1.0\textwidth]{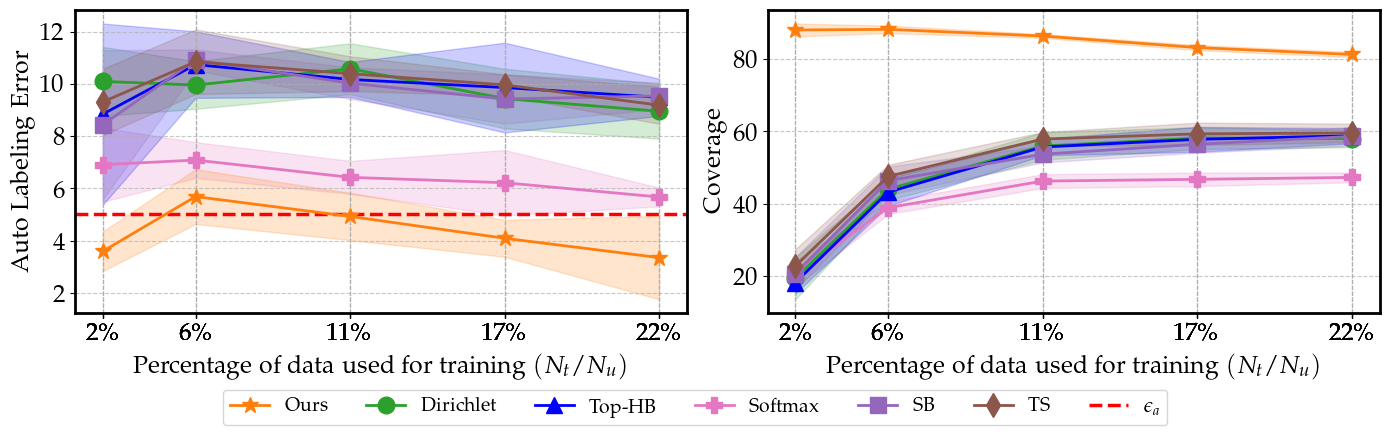}
    \caption{\normalsize Auto-labeling error and coverage for different post-hoc methods on 20 Newsgroups while we vary $N_t$. $N_u=9,052$ is the size of the given unlabeled pool.}
    \label{fig:nt-20newsgroups}
  \end{subfigure}  

  \vspace{1em}
  \label{fig:varying-nt-graphs}
\end{figure*}

\begin{figure*}[h]
  \centering
  \begin{subfigure}
    \centering
    \includegraphics[width=1.0 \textwidth]{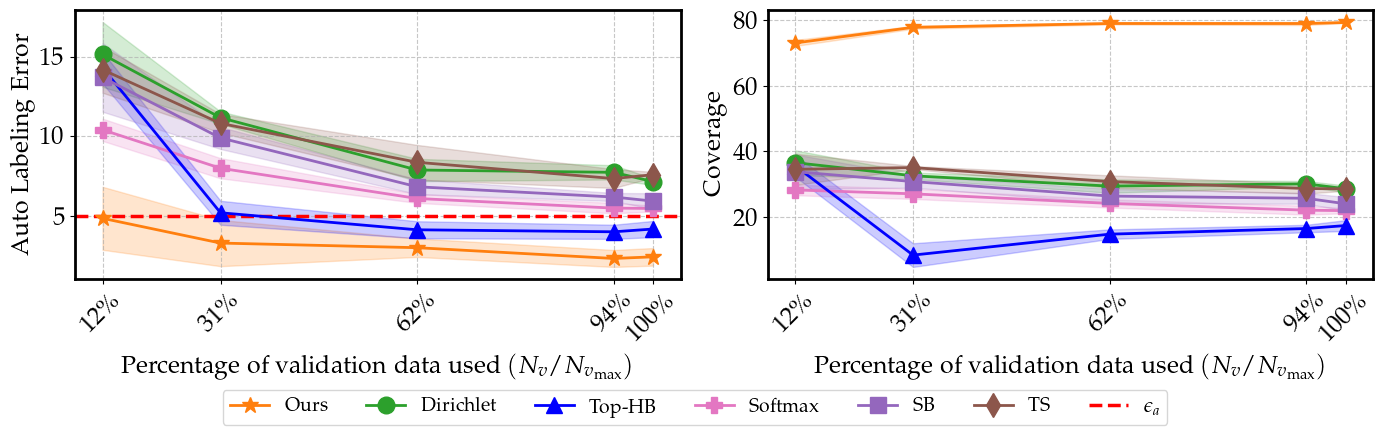}
    \caption{\normalsize Auto-labeling error and coverage for different post-hoc methods on CIFAR-10 while we vary $N_v$. $N_{v_\mathrm{max}}=8,000$ is the maximum number of points available for validation.}
    \label{fig:nv-cifar10}
  \end{subfigure}  
  
  \begin{subfigure}
    \centering
    \includegraphics[width=1.0 \textwidth]{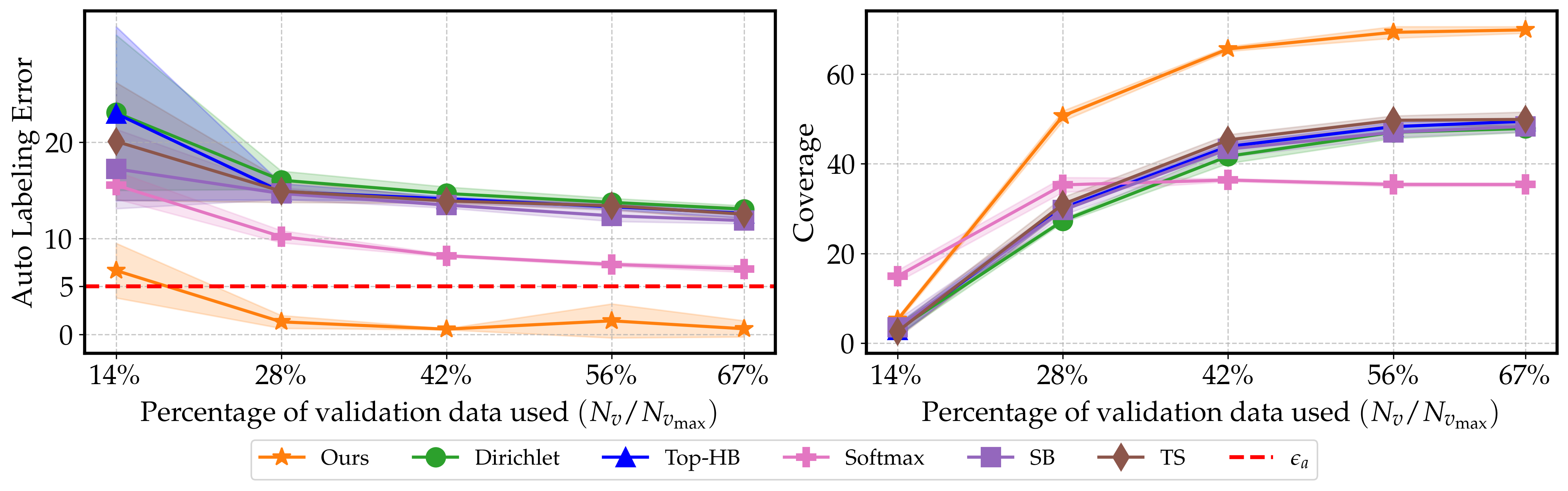}
    \caption{\normalsize Auto-labeling error and coverage for different post-hoc methods on Tiny-ImageNet while we vary $N_v$. $N_{v_\mathrm{max}}=18,000$ is the maximum number of points available for validation.}
    \label{fig:nv-tinyimagenet}
  \end{subfigure}  
  
  \begin{subfigure}
    \centering
    \includegraphics[width=1.0 \textwidth]{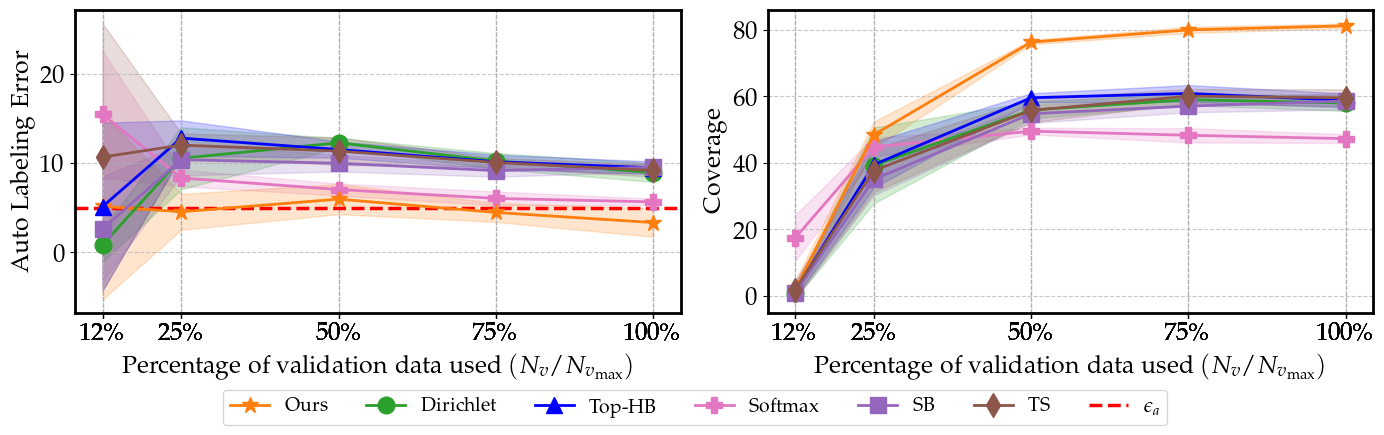}
    \caption{\normalsize Auto-labeling error and coverage for different post-hoc methods on 20 Newsgroups while we vary $N_v$. $N_{v_\mathrm{max}}=1,600$ is the maximum number of points available for validation.}
    \label{fig:nv-20newsgroups}
  \end{subfigure}  
  \vspace{1em}
  \label{fig:varying-nv-graphs}
\end{figure*}

\begin{table}[h]
\centering
\begin{tabular}{@{}llll@{}}
\toprule
\textbf{Feature} & \textbf{Model} & \textbf{Error} & \textbf{Coverage} \\ \midrule
Pre-logits       & Two Layer      & 4.6 $\pm$ 0.3  & 82.8 $\pm$ 0.5    \\
Logits           & Two Layer      & 3.2 $\pm$ 1.3  & 82.8 $\pm$ 0.3    \\
Concat           & Two Layer      & 3.3 $\pm$ 0.8  & 82.9 $\pm$ 0.4    \\ \bottomrule
\end{tabular}
\caption{Auto-labeling error and coverage for the 3 feature representations we could use for 20 Newsgroup. As we can see, the feature representation does not lead to a significant difference in auto-labeling error and coverage.} \label{table:features_20news}
\end{table}

\begin{table}[h]
\centering
\begin{tabular}{@{}llll@{}}
\toprule
\textbf{Feature} & \textbf{Model} & \textbf{Error} & \textbf{Coverage} \\ \midrule
Pre-logits       & Two Layer      & 2.1 $\pm$ 0.5  & 79.0 $\pm$ 0.2    \\
Logits           & Two Layer      & 3.1 $\pm$ 0.4  & 76.5 $\pm$ 0.9    \\
Concat           & Two Layer      & 2.3 $\pm$ 0.5  & 79.0 $\pm$ 0.3    \\ \bottomrule
\end{tabular}
\caption{Auto-labeling error and coverage for the 3 feature representations we could use for CIFAR10 SimpleCNN. As we can see, the feature representation does not lead to a significant difference in auto-labeling error and coverage.}
\label{table:cifar10}
\end{table}

\subsection{Experiments on \ourmethod~input}

Figure \ref{fig:illu_our} illustrates that we could use logits (last layer's representations), pre-logits (second last layer's representations), or the concatenation of these two as the input to $\fg$. To help us decide which one we should use, we conduct a hyperparameter search for input features on the CIFAR-10 and 20 Newsgroup dataset using the Squentropy train-time method. Table \ref{table:features_20news} and \ref{table:cifar10} present the auto-labeling error and coverage of using the 3 types of feature representations. As we can see, all feature
representation leads to a similar auto-labeling error and coverage, and in some cases, it is better to include pre-logits as well. Therefore, we use concatenated representation (Concat), allowing more flexibility.



\subsection{Hyperparameters}
\label{subsec:hyperparameters}
The hyperparameters and their values we swept over are listed in Table \ref{table:train_time_hyp_params} and \ref{table:post_hoc_hyp_params} for train-time and post-hoc methods, respectively.
\begin{table}[]
\centering
\begin{tabular}{@{}lll@{}}
\toprule
\textbf{Method}                  & \textbf{Hyperparameter} & \textbf{Values}           \\ \midrule
\multirow{6}{*}{Common} & optimizer      & SGD              \\
                        & learning rate  & 0.001, 0.01, 0.1 \\
                        & batch size     & 32, \underline{256}            \\
                        & max epoch      & 50, \underline{100}            \\
                        & weight decay   & 0.001, 0.01, 0.1 \\
                        & momentum       & 0.9              \\ \midrule
\multirow{2}{*}{CRL}    & rank target    & softmax          \\
                        & rank weight    & 0.7, 0.8, 0.9    \\ \midrule
FMFP                    & optimizer      & SAM              \\ \bottomrule
\end{tabular}
\caption{Hyperparameters swept over for train-time methods.  Those listed next to Common are the hyperparameters for the four train-time methods: Vanilla, CRL, FMFP, and Squentropy. Therefore, we do not list those again for each method. Note that for FMFP, we used SAM optimizer instead of SGD. For each method, we swept through all possible combinations of the possible values for each hyperparameter. Underlined values are only used on TinyImageNet since it is a complicated dataset containing 200 classes.}
\label{table:train_time_hyp_params}
\end{table}

\begin{table}[b]
\centering
\begin{tabular}{@{}lll@{}}
\toprule
\textbf{Method}                      & \textbf{Hyperparameter}           & \textbf{Values}            \\ \midrule
Temperature scaling         & optimizer                & Adam             \\
                            & learning rate            & 0.001, 0.01, 0.1 \\
                            & batch size               & 64               \\
                            & max epoch                & 500              \\
                            & weight decay             & 0.01, 0.1, 1     \\ \midrule
Top-label histogram binning & points per bin           & 25, 50           \\ \midrule
Scaling-binning             & number of bins           & 15, 25           \\
                            & learning rate            & 0.001, 0.01, 0.1 \\
                            & batch size               & 64               \\
                            & max epoch                & 500              \\
                            & weight decay             & 0.01, 0.1, 1     \\ \midrule
Dirichlet calibration       & regularization parameter & 0.001, 0.01, 0.1 \\ \midrule
Ours   & $\lambda$           & 10, 100      \\
       & features key   & concat       \\
       & class-wise     & independent  \\
       & optimizer      & Adam         \\
       & learning rate  & 0.01, 0.1    \\
       & max epoch      & 500          \\
       & weight decay   & 0.01, 0.1, 1 \\
       & batch size     & 64           \\
       & regularize     & false        \\
       & $\alpha$       & 0.01, 0.1, 1 \\ \bottomrule
\end{tabular}
\caption{Hyperparamters swept over for post-hoc methods. For each method, we swept through all possible combinations of the possible values for each hyperparameter. }
\label{table:post_hoc_hyp_params}
\end{table}

\subsection{Train-time and post-hoc methods} \label{sec:train-time-post-hoc}
    \subsubsection{Train-time methods}
        \begin{enumerate}
            \item \textit{Vanilla}: Neural networks are commonly trained by minimizing the cross entropy loss using stochastic gradient descent (SGD) with momentum ~\citep{amari1993backpropagation, Bottou2012}. We refer to this as the Vanilla training method. We also include weight decay to mitigate the overconfidence issue associated with this method ~\citep{guo2017calibration}.
            
            \item \textit{Squentropy} \citep{hui2023Squentropy}:  This method adds the average square loss over the incorrect classes to the cross-entropy loss. This simple modification to the Vanilla method leads to the end model with better test accuracy and calibration.
            
            \item \textit{Correctness Ranking Loss (CRL)} \citep{moon2020CRL}: This method includes a term in the loss function of the vanilla training method so that the confidence scores of the model are aligned with the ordinal rankings criterion ~\citep{hendrycks2017OrdinalOOD, corbi2019Confidnet}. The confidence functions satisfying this criterion produce high scores on points where the probability of correctness is high and low scores on points with low probabilities of being correct.
            
            \item \textit{FMFP} ~\citep{zhu2022FMFP} aims to align confidence scores with the ordinal rankings criterion. It uses Sharpness Aware Minimizer (SAM) \citep{foret2021sharpnessaware} to train the model, with the expectation that the flat minima would benefit the ordinal rankings objective of the confidence function.
        \end{enumerate}
        
\subsubsection{Post-hoc methods}
    \begin{enumerate}
        \item \textit{Temperature scaling} \citep{guo2017calibration}: This is a variant of Platt scaling ~\citep{guo2017calibration}, a classic and one of the easiest parametric methods for post-hoc calibration. It rescales the logits by a learnable scalar parameter and has been shown to work well for neural networks.
        
        \item \textit{Top-Label Histogram-Binning} ~\citep{gupta2021top}: Since TBAL assigns the top labels (predicted labels) to the selected unlabeled points, it is appealing to only calibrate the scores of the predicted label. Building upon a rich line of histogram-binning methods (non-parametric) for post-hoc calibration \citep{zadrozny2002transforming}, this method focuses on calibrating the scores of predicted labels.
    
        \item \textit{Scaling-Binning} ~\citep{kumar2019verified}: This method combines parametric and non-parametric methods. It first applies temperature scaling and then bins the confidence function values to ensure calibration. 
    
        \item \textit{Dirichlet Calibration} ~\citep{kull2019Dirichlet}: This method models the distribution of predicted probability vectors separately on instances of each class and assumes the class conditional distributions are Dirichlet distributions with different parameters. It uses linear parameterization for the distributions, which allows easy implementation in neural networks as additional layers and softmax output.
    \end{enumerate}

\textbf{Note:} For binning methods, uniform mass binning \citep{zadrozny2002transforming} has been a better choice over uniform width binning. Hence, we use uniform mass binning as well.

\subsection{Compute resources}
Our experiments were conducted on machines equipped with the NVIDIA RTX A6000 and NVIDIA GeForce RTX 4090 GPUs.

\subsection{Detailed dataset and model}
    \begin{enumerate}
        \item The MNIST dataset \cite{lecun1998mnist} consists of $28 \times 28$ grayscale images of hand-written digits across 10 classes. It was used alongside the LeNet5 \cite{lecun1998LeNet}, a convolutional neural network, for auto-labeling.
        
        \item The CIFAR-10 dataset \cite{krizhevsky2009CIFAR-10} contains $3 \times 32 \times 32$ color images across 10 classes. We utilized its raw pixel matrix in conjunction with SimpleCNN \cite{mednet}, a convolutional neural network with approximately 5.8M parameters, for auto-labeling.
        
        \item Tiny-ImageNet \cite{le2015tiny} is a color image dataset that consists of 100K images across 200 classes. Instead of using the $3 \times 64 \times 64$ raw pixel matrices as input, we utilized CLIP \cite{radford2021learning} to derive embeddings within the $\mathbb{R}^{512}$ vector space. We used a 3-layer perceptron (1,000-500-300) as the auto-labeling model. 
        \item 20 Newsgroups \cite{misc_twenty_newsgroups_113, scikit-learn} is a natural language dataset comprising around 18,000 news posts across 20 topics. We used the FlagEmbedding \cite{bge_embedding} to map the textual data into $\mathbb{R}^{1024}$ embeddings. We used a 3-layer perceptron (1,000-500-30) as the auto-labeling model.

    \end{enumerate}

\subsection{Detailed experiments protocol}
\label{subsec:exp-protocol}

We predefined TBAL hyperparameters for each dataset-model pair and the hyperparameters we will sweep for each train-time and post-hoc method in Table \ref{table:train_time_hyp_params} and Table \ref{table:post_hoc_hyp_params} respectively. For a dataset-model pair, initially, we perform a hyperparameter search for the train-time method. Subsequently, we optimize the hyperparameters for post-hoc methods while keeping the train-time method fixed with the previously found optimum hyperparameter for that dataset-model pair.

We fix the hyperparameters for the train-time method while searching hyperparameters for the post-hoc method to alleviate computational budget throttle. We effectively reduce the search space to the sum of the cardinalities of unique hyper-parameter combinations across the two methods instead of a larger multiplicative product. Furthermore, due to the independent nature of these hyper-parameter combinations, TBAL runs can be highly parallelized to expedite the search process.

Since TBAL operates iteratively to acquire human labels for model training, selecting hyper-parameters at each round of TBAL could quickly become intractable and lose its practical significance. To better align with its practical usage, we only conducted a hyperparameter search for the initial TBAL round. The specific set of hyperparameters used for the search are reported in Table \ref{table:post_hoc_hyp_params}. 

After completing the hyperparameter search for train-time and post-hoc methods, the determined hyperparameter combinations are subjected to a full evaluation across all iterations of TBAL. At the end of each iteration, the auto-labeled points are evaluated against their ground truth labels to determine their auto-labeling error. These points are then added to the auto-labeled set, where their ratio to the total amount of unlabeled data determines the coverage. This iterative process continues until all unlabeled data are exhaustively labeled by either the oracle or through auto-labeling in the final iteration. The auto-labeling error and coverage at the final iteration of TBAL are then recorded. 

Since TBAL incorporates randomized components as detailed in Algorithm \ref{alg:main_algo}, we ran the algorithm 5 times, each with a unique random seed while maintaining the same hyperparameter combination. We then recorded the results from the final iteration of these runs and calculated the mean and standard deviation of both auto-labeling error and coverage. These figures are reported in Table \ref{table:final_table}.


A limitation of the grid search approach in hyper-parameter optimization becomes apparent when our predefined hyper-parameter choices result in sub-optimal coverage and auto-labeling errors. Using these sub-optimal hyper-parameters can adversely affect the multi-round iterative process in TBAL, prompting the need for repetitive searches to find more effective hyper-parameters. When encountering such scenarios, TBAL users should explore additional hyper-parameter options until satisfactory performance is achieved in the initial round. However, we opted for a more straightforward approach to hyper-parameter selection, mindful of the computational demands of repeatedly optimizing multiple hyper-parameters across different methods. In scenarios expressed conditionally, we retained the top-1 hyper-parameter combination for any given method if it achieved the highest coverage while adhering to the specified error margin ($\epsilon_a$). If no hyper-parameter combinations yielded an auto-labeling error at most equal to the error margin ($\epsilon_a$), we then chose the hyper-parameter combination with the lowest auto-labeling error, regardless of its coverage. In the case of ties, we resolved them through random selection. This process results in obtaining singular values for each choice of hyper-parameter after completing each method's hyper-parameter search.

\end{document}